\documentclass[letterpaper, 10 pt, conference]{ieeeconf}
\IEEEoverridecommandlockouts
\usepackage[utf8]{inputenc}
\usepackage[T1]{fontenc}
\usepackage[hidelinks]{hyperref}
\usepackage{url}
\usepackage{booktabs}
\usepackage{amsmath}
\usepackage{amsfonts}
\usepackage{microtype}
\usepackage{xcolor}
\usepackage{graphicx}
\let\labelindent\relax
\usepackage{enumitem}
\usepackage{orcidlink}

\newcommand{\orcid}[1]{\,\orcidlink{#1}}

\newlength{\qualw}
\newcommand{\qpanel}[1]{\includegraphics[width=\qualw]{#1}}
\newcommand{\qhead}[1]{\scriptsize #1}

\definecolor{savegreen}{rgb}{0.00,0.45,0.10}
\newcommand{\memsave}[1]{\textcolor{savegreen}{\textbf{#1}}}

\newcommand{\rhead}[1]{\medskip\noindent\textbf{#1.}\ \ignorespaces}

\title{\LARGE \bf Toward On-Chip Training of Spiking Neural Networks for Dense Event-Based Vision}

\author{\authorblockN{Maxime Vaillant$^{1,2,4}$\orcid{0009-0006-5948-948X},
Axel Carlier$^{1,3}$\orcid{0000-0002-6838-3445},
Lai Xing Ng$^{1,4}$\orcid{0000-0002-5457-6289},\\
Christophe Hurter$^{1,3}$\orcid{0000-0003-4318-6717}
and Benoit R. Cottereau$^{1,5}$\orcid{0000-0002-2624-7680}\vspace{0.6ex}}
\authorblockA{$^1$CNRS, IPAL IRL 2955, Singapore\\
$^2$Universit\'e de Toulouse, IRIT, France\\
$^3$F\'ed\'eration ENAC ISAE-SUPAERO ONERA, Universit\'e de Toulouse, France\\
$^4$Institute for Infocomm Research, A*STAR, Singapore\\
$^5$CerCo, CNRS UMR 5549, Universit\'e de Toulouse, France\\[0.5ex]
maxime.vaillant@utoulouse.fr, axel.carlier@isae-supaero.fr, ng\_lai\_xing@a-star.edu.sg,\\
christophe.hurter@enac.fr, benoit.cottereau@cnrs.fr}
}

\begin{document}

\maketitle
\begin{abstract}
Event cameras provide low-latency, asynchronous visual sensing for resource-constrained robotics. Spiking neural networks (SNNs) process event streams naturally, but training deep SNNs with backpropagation through time (BPTT) requires substantial memory and remains difficult on neuromorphic hardware. Local learning avoids this by restricting error propagation to local blocks, but existing methods mainly target classification rather than dense prediction. We introduce DELL (Dense Event-driven Local Learning), a block-wise scheme for dense event-based vision that replaces global gradient propagation with local dense supervision. Learnable, spatially structured local heads supervise each block at its appropriate resolution while preserving temporal dynamics within blocks. We evaluate DELL on optical-flow regression and semantic segmentation with a fully spiking U-shaped architecture. On DSEC optical flow, DELL reduces peak training memory by 39.6\% relative to end-to-end BPTT while improving accuracy, reaching 1.670 px endpoint error on the official test benchmark versus 1.941 px for the same backbone trained end-to-end. Block detachment behaves more like a regularizer than a constraint. DECOLLE, the existing local-learning baseline, relies on fixed random local read-outs poorly suited to dense regression, resulting in a 3.9× higher endpoint error; learnable local heads recover this loss and outperform end-to-end training across all optical-flow metrics. On segmentation, they recover most of the performance gap, although DELL remains a few mIoU points behind end-to-end training. With 2.3M parameters, 24× fewer than the strongest SNN baseline, the backbone remains competitive with the SNN state of the art on DSEC. These results extend local learning to dense event-based prediction while substantially reducing training memory.

\end{abstract}

\section{Introduction}
\label{sec:introduction}

Event cameras provide an attractive sensing modality for robotic systems by encoding visual information as asynchronous streams of per-pixel brightness changes. Their sparse output and microsecond temporal resolution are particularly well suited to scenes with fast dynamics, while potentially reducing data transfer and computational requirements~\cite{gallego2022survey}. Spiking neural networks (SNNs) naturally match this representation, processing asynchronous spikes over time without requiring the reconstruction of dense video frames. Together, event cameras and SNNs therefore suit low-latency, resource-constrained robotic perception.

On-chip training is an important step toward fully exploiting these advantages in autonomous robots. Neuromorphic processors such as Loihi~\cite{davies2018loihi} provide a promising platform for deploying and training SNNs, but standard backpropagation through time (BPTT) is difficult to implement efficiently on such hardware. BPTT requires maintaining the computational graph across both network depth and time and propagating error signals through the entire network~\cite{roy2019spike}, which is costly in memory and non-local. Reducing this reliance is therefore a direction toward memory-efficient, hardware-friendly training.

\begin{figure*}[t]
\centering
\includegraphics[width=\textwidth]{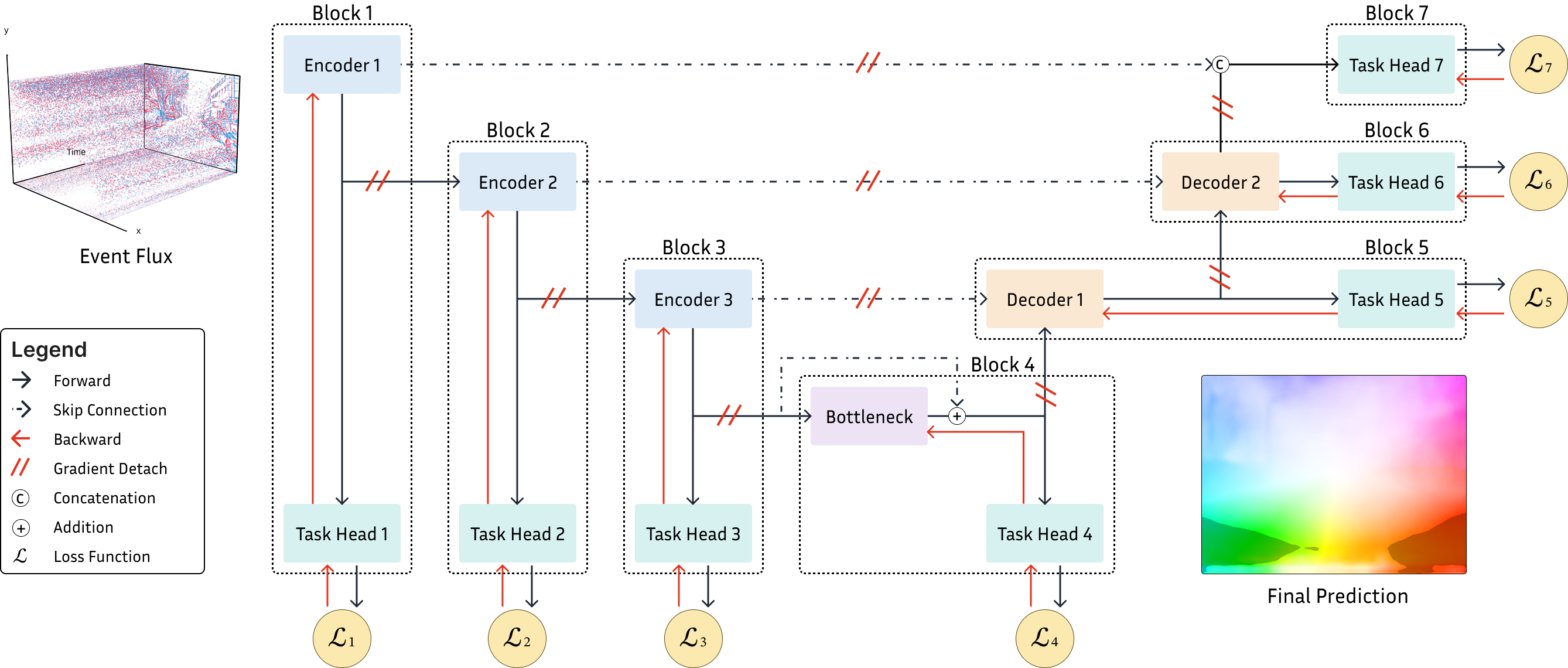}
\caption{Block-wise local learning: each block has its own local head and local loss $\mathcal{L}_i$, with gradients detached at every block boundary (symbols per the inset legend). In \texttt{DELL}, encoder and bottleneck blocks reach their local head through a training-only decoder module that restores the resolution of the block's input. Only the decoder and read-out heads' predictions are combined into the final output map at inference; encoder and bottleneck local heads are training-only. Shown with a reduced block count for legibility; the full network uses 11 blocks (Section~\ref{sec:blockwise}).}
\label{fig:blockwise}
\end{figure*}

Block-wise local learning restricts error propagation to local portions of the network. \texttt{DECOLLE}~\cite{kaiser2020decolle} detaches gradients at layer boundaries and trains each layer using a local loss and a fixed, random read-out. Subsequent approaches have replaced these read-outs with learnable local modules~\cite{guo2023truncatedbptt,ma2025stdl}, while related work has investigated local error signals in non-spiking networks~\cite{nokland2019local,belilovsky2019greedy}. However, these approaches have primarily been developed and evaluated for classification, where a single global label can supervise local predictions at different depths. Extending local learning to dense robotic perception therefore raises a different challenge: intermediate representations must predict spatially structured targets, while local predictions may be produced at different spatial resolutions.

Here, we address this challenge with \texttt{DELL} (\emph{Dense Event-driven Local Learning}), which trains a dense event-based prediction network one block at a time (Figure~\ref{fig:blockwise}). Rather than propagating a global error signal through the network, \texttt{DELL} provides each block with a local dense target and detaches gradients between blocks, allowing the blocks to be optimized independently while preserving their temporal dynamics. We evaluate \texttt{DELL} on two complementary dense prediction tasks: optical flow on DSEC~\cite{gehrig2021dsec}, a per-pixel regression problem relevant to ego-motion estimation, obstacle avoidance, and motion-aware navigation, and semantic segmentation on M3ED~\cite{chaney2023m3ed}, a per-pixel classification problem relevant to scene understanding. Using the same compact 2.3M-parameter U-shaped SNN backbone, designed under the neuromorphic inference constraints of Section~\ref{sec:design-constraints}, we compare \texttt{DELL} with standard end-to-end BPTT (\texttt{E2E}) and \texttt{DECOLLE}-style training with fixed random local heads (\texttt{s-DECOLLE}). This lets us quantify the accuracy and memory effects of local dense supervision and assess how the choice of local read-outs affects dense prediction. Section~\ref{sec:expflow} compares our backbone with the SNN state of the art on DSEC, then end-to-end against block-wise training in accuracy and peak training memory across depths. Section~\ref{sec:segmentation} covers semantic segmentation.

This paper makes the following contributions:
\begin{itemize}
\item We introduce \texttt{DELL} (\emph{Dense Event-driven Local Learning}), a block-wise learning scheme for dense event-based prediction that replaces gradient propagation across network blocks with local dense supervision. On DSEC optical flow, \texttt{DELL} reduces peak training memory by 39.6\% while improving accuracy over end-to-end training; on semantic segmentation it remains within a few mIoU points of end-to-end training.

\item We establish experimentally that fixed random local read-outs are poorly suited to dense prediction, whereas learnable spatially structured local heads recover this degradation and can outperform end-to-end training.

\item We develop a compact, fully spiking backbone for dense event-based prediction, adapting the U-shaped architecture of Cuadrado et al.~\cite{cuadrado2023} to a strictly streaming setting. With 2.3M parameters, 24$\times$ fewer than the strongest SNN baseline, it remains competitive with the SNN state of the art on DSEC when trained with \texttt{DELL}.
\end{itemize}

Code is available at \url{https://github.com/maxime-vaillant/DELL}.

\section{Related Work}
\label{sec:related}

\rhead{Neuromorphic hardware and local learning} Digital neuromorphic accelerators such as Loihi \cite{davies2018loihi} are designed around sparse, event-driven computation and local synaptic learning based on pre- and post-synaptic activity. This locality contrasts with standard backpropagation through time (BPTT), which requires error signals to be propagated across network depth and time \cite{roy2019spike}. Block-wise local learning provides an alternative by detaching gradients at block boundaries, thereby restricting credit assignment to local signals. DECOLLE \cite{kaiser2020decolle} was among the first approaches to combine inter-layer gradient detachment with local learning in SNNs. Each layer is equipped with a fixed, random read-out and trained online using eligibility traces rather than BPTT. DECOLLE was demonstrated on small classification benchmarks with shallow networks.
More recent approaches use learnable local classifiers. Guo et al. \cite{guo2023truncatedbptt} detach gradients at block boundaries while learning local classifiers, recovering much of the accuracy lost by temporally truncated BPTT on image classification. Ma et al. \cite{ma2025stdl} similarly decouple spatial and temporal credit assignment using learnable local modules. However, these approaches are evaluated on classification tasks, where a global label can be provided to each local module. Dense prediction instead requires local targets that preserve the spatial structure of the output and remain meaningful throughout the temporal dynamics of the SNN. Related local-error strategies have also been explored in non-spiking networks. Nøkland and Eidnes \cite{nokland2019local} trained deep convolutional networks using purely local error signals, while Belilovsky et al. \cite{belilovsky2019greedy} demonstrated greedy layer-wise supervised training at ImageNet scale. These studies show that local learning can approach end-to-end performance while avoiding gradient propagation across layer boundaries. Building on this principle, DELL introduces learnable decoder-style local heads specifically designed for dense spatiotemporal prediction from event streams.

\rhead{SNNs for event-based optical flow} Several recent works have investigated spiking architectures for optical flow on the DSEC benchmark, providing the main points of comparison for our approach \cite{cuadrado2023,kosta2023adaptivespikenet,tian2024sdformerflow}. Cuadrado et al. \cite{cuadrado2023} proposed a U-Net-shaped SNN that aggregates temporal information through Conv3d layers fused with each convolutional stage, rather than relying solely on LIF membrane recurrence. It is trained end-to-end with full BPTT over the input window.

Adaptive-SpikeNet \cite{kosta2023adaptivespikenet} uses a fully event-driven, timestep-recurrent LIF backbone with learnable neuronal dynamics instead of fixed leak and threshold parameters. It is evaluated on MVSEC and DSEC using surrogate-gradient BPTT and reports substantial parameter and energy savings compared with comparably accurate ANNs. Like Cuadrado et al., however, it relies on end-to-end training without depth-wise gradient detachment. Its DSEC figures are not reported on the official test benchmark, unlike those of Cuadrado et al.\ and SDformerFlow, and it is therefore absent from Table~\ref{tab:sota}; its efficiency analysis likewise covers MVSEC alone.

More recently, SDformerFlow \cite{tian2024sdformerflow} introduced a fully spiking Spikeformer based on spatiotemporal shifted-window self-attention. It reports the best performance among the SNN approaches considered here and, to our knowledge, represents the current state of the art for SNN optical flow on DSEC. However, its architecture is substantially larger than ours, with approximately 24$\times$ more parameters, and it is likewise trained end-to-end without a local or block-wise objective.

Overall, existing SNN approaches to event-based optical flow rely on end-to-end BPTT, leaving block-wise local training for dense spatiotemporal prediction largely unexplored.

\rhead{SNNs for event-based semantic segmentation} SpikingEDN \cite{zhang2024spikingedn} uses an adaptive-threshold LIF encoder-decoder for sparse event inputs. On DSEC-Semantic, its events-only LIF configuration achieves 52.71\% mIoU, providing a direct comparison with our setting. EvSegSNN \cite{hareb2024evsegsnn} instead proposes a lightweight, fully spiking U-Net with parametric LIF neurons, reporting a 5.58-point absolute mIoU improvement over its baseline on DDD17 while using 62\% fewer parameters.

These works rely on end-to-end training and do not investigate block-wise local credit assignment. In contrast, our study evaluates local and end-to-end training on DSEC-Semantic, in addition to optical flow, and further considers M3ED. To our knowledge, M3ED has not previously been evaluated with an SNN for semantic segmentation; our results therefore provide a first reference point for spiking event-based segmentation on this dataset.

Overall, existing SNN approaches to event-based dense prediction rely predominantly on end-to-end BPTT. While local or block-wise learning has been explored for SNN classification, its application to dense spatiotemporal prediction remains largely unexplored. DELL addresses this gap by introducing learnable decoder-style local heads that provide spatially structured targets at intermediate blocks, enabling local credit assignment for event-based dense prediction.

\section{Method}
\label{sec:method}

\subsection{Network Architecture}
\label{sec:design-constraints}
\label{sec:architecture}

Our network follows a U-shaped encoder-bottleneck-decoder architecture inspired by Cuadrado et al.~\cite{cuadrado2023}. We adapt this design to a strictly streaming, event-driven setting where temporal state comes only from leaky integrate-and-fire (LIF) neurons, whose membrane potential $u_i^{(l)}[t]$ at layer $l$ follows the discrete-time update rule
\begin{equation}
    \label{eq:lif}
u_i^{(l)}[t] = \beta u_i^{(l)}[t-1] \big(1 - s_i^{(l)}[t-1]\big) + \sum_j w_{ij} s_j^{(l-1)}[t],
\end{equation}
where $\beta \in (0,1)$ is the membrane-leak decay factor, $w_{ij}$ the synaptic weight from neuron $j$ in the previous layer to neuron $i$, and $s_j^{(l-1)}[t] \in \{0, 1\}$ the input spike. A neuron emits a spike $s_i^{(l)}[t]$ when its membrane potential exceeds a firing threshold $V_{th}$:
\begin{equation}
    \label{eq:heaviside}
s_i^{(l)}[t] = \Theta(u_i^{(l)}[t] - V_{th}),
\end{equation}
where $\Theta(\cdot)$ is the Heaviside step function. The factor $\big(1 - s_i^{(l)}[t-1]\big)$ in \eqref{eq:lif} implements a hard reset with a reset potential of zero. The decay factor is tied to the membrane time constant by $\beta = 1 - 1/\tau$, with $\tau = 3.2$ network-wide. Since $\Theta$ is not differentiable, we use the ATan surrogate gradient.

The only temporal operator in the network is the LIF cell's own membrane recurrence: the forward pass processes one timestep at a time and produces a prediction before the next arrives. This streaming constraint distinguishes our architecture from the Conv3d temporal-fusion layers used by Cuadrado et al.~\cite{cuadrado2023}. The backbone uses no batch normalization, Conv3d, transposed convolutions, attention, or softmax, and uses only operations compatible with digital neuromorphic inference.

The backbone uses $\text{base\_channels}{=}32$, with five encoder stages of widths $32{-}64{-}128{-}256{-}512$. Each stage applies a convolution, the LIF update and max-pooling, except level $0$, which stays at full spatial resolution. The bottleneck is a two-convolution SEW residual block~\cite{fang2021sew} of width $512$. Each of the four decoder blocks performs nearest-neighbor upsampling, concatenates the encoder skip connection, and applies a convolution followed by a LIF update. Following Cuadrado et al.~\cite{cuadrado2023}, we use separable convolutions in the deepest encoder stage, the first decoder block and the bottleneck, and standard Conv2d elsewhere.

The read-out is a $3{\times}3$ convolution producing $C$ channels, the only task-dependent component of the network: $C{=}2$ for the $(u,v)$ components of optical flow, and $C$ equal to the number of classes for semantic segmentation. Predictions are averaged over the $T$ input timesteps, yielding a dense per-pixel map at backbone resolution. The backbone is partitioned into blocks for block-wise local learning (Section~\ref{sec:blockwise}). All three training configurations share the same backbone and the same inference parameter count, 2.3M on optical flow; the auxiliary local-learning components are used only during training, discarded at inference.
\subsection{Block-wise Local Learning}
\label{sec:blockwise}

We treat each encoder stage, the bottleneck, each decoder stage, and the read-out as one block ($B{=}11$ blocks total, Figure~\ref{fig:blockwise}). Every prediction, at the final output and at every local head, is scored with the same task loss, the only part of the scheme that changes between tasks. On optical flow we use a modulus-angular loss, an equally-weighted sum of an endpoint-error term and an angular term, following Cuadrado et al.~\cite{cuadrado2023}. On semantic segmentation, the read-out and every local head produce 11 per-pixel class scores, trained with a SmoothL1 loss against the one-hot encoded label map. We also tried cross-entropy and Dice losses, and retained SmoothL1, which we found better behaved in the spiking setting. In both cases the loss is averaged only over the pixels carrying a valid target, rather than over the whole frame; DSEC flow ground truth in particular is sparse.

In \texttt{s-DECOLLE} and \texttt{DELL}, gradients are detached at every block boundary, so a block's gradient reaches only its own parameters. Each block is trained, with its own optimizer, against its own local loss, through its own local head, architecturally identical to the final read-out head. A block at a coarser resolution is supplied a target at its own resolution by Mask Average Pooling, on both tasks: within each pooling window, only pixels carrying valid ground truth are averaged, rather than the whole window, so invalid pixels do not bias the pooled target. On segmentation this pooling is applied channel-wise to the one-hot label map, so the pooled target is soft rather than a class index: each channel holds the fraction of valid pixels of that class inside the window. In \texttt{DELL}, encoder and bottleneck blocks additionally pass their output through a training-only decoder module before their local head: each encoder stage halves the spatial dimensions, and this module restores them, with nearest-neighbor upsampling by a factor of two, concatenation of the corresponding encoder skip connection, then a convolution followed by a LIF update, the same composition as a decoder block (Section~\ref{sec:architecture}). This local loss is therefore computed at the resolution of the block's own input, not the pooled resolution of its output. These modules account for the 5.6\,M difference between \texttt{DELL}'s training-time and inference parameter counts (Table~\ref{tab:tradeoff}).

Detaching gradients at block boundaries also changes what the backward pass must keep in memory. End-to-end BPTT must retain, simultaneously, the activations of all $B$ blocks across all $T$ timesteps: an activation footprint of $\mathcal{O}(B\cdot T)$. Under block-wise detachment, at most one block's activations need be resident at a time, since each block's backward pass consumes and releases its own activations before the next block's forward pass begins, reducing the footprint to $\mathcal{O}(\max_i|b_i|\cdot T)$, where $|b_i|$ denotes the size of block $i$'s own activations. This asymptotic reduction is the reason for the peak-memory reduction measured in Section~\ref{sec:expflow} (39.6\% at depth 4); the two figures need not coincide numerically, since Table~\ref{tab:tradeoff}'s measurement also reflects optimizer-state and implementation overheads outside this simplified accounting.

We compare three configurations sharing this backbone. \texttt{E2E} is the end-to-end baseline: no detachment, a single global loss backpropagated through depth and time. \texttt{s-DECOLLE} uses the same detachment schedule and local losses, but every local head other than the read-out, the five encoder heads and the bottleneck head, is frozen at its random initialization. The read-out head remains learnable, trained against its own local loss like every other block. \texttt{DELL} (ours) trains every head jointly with its block and equips the encoder and bottleneck heads with the decoder-style module described above.

At inference, every local head other than the read-out is discarded, along with \texttt{DELL}'s prepended decoder-style modules. What remains are the four decoder blocks' own predictions and the read-out's prediction: five predictions at four distinct resolutions, since the final decoder block and the read-out share the backbone resolution of $192{\times}256$. Each prediction is bilinearly upsampled to that backbone resolution and averaged into a single, smooth output map, which acts as a low-pass filter that the two tasks react to differently. Optical flow is a spatially smooth field, scored per pixel by magnitude, so the coarse scales contribute useful global structure, whereas semantic segmentation is scored by mIoU, an unweighted mean over classes dominated by small and thin classes, which the same smoothing removes. For this reason the multi-scale combination applies to optical flow only, across all three configurations; on segmentation every configuration's reported output is the read-out prediction alone, and \texttt{E2E} is correspondingly trained with the read-out loss alone. \texttt{DELL} and \texttt{s-DECOLLE} still train every block, including the encoder and bottleneck, against its own local loss on segmentation. On optical flow, this combined prediction is then further upsampled to DSEC's native $480{\times}640$ grid and magnitude-rescaled by $2.5{\times}$ for scoring (Section~\ref{sec:expflow}). On this task the output pipeline is therefore identical across the three configurations, which share the same inference network and the same inference parameter count of 2.3M, despite differing training-time parameter counts (Table~\ref{tab:tradeoff}).
\section{Experiments}
\label{sec:experiments}

\subsection{Optical Flow}
\label{sec:expflow}

We first test the per-pixel \emph{regression} setting, optical flow, where no single global label exists per input.

\rhead{Setup}\label{sec:setup}

We use DSEC~\cite{gehrig2021dsec}, with the same 13-training/5-validation recording split (18 recordings total) as Cuadrado et al.~\cite{cuadrado2023} and SDformerFlow~\cite{tian2024sdformerflow}. Each sample is a $\sim$100\,ms causal window of events from the rectified left camera, binned into $T{=}16$ time steps. These bins are processed in strictly causal order, with no access to future bins, and the resulting tensor is downsampled to $192{\times}256$. We report endpoint error (EPE, px), angular error (AE, deg), and outlier rate (\%, EPE${}>{}$3px). All models are trained with AdamW under a cosine annealing schedule, with the same hyperparameters and budget across the three configurations. Training uses full FP32 precision, without gradient accumulation or gradient checkpointing. The memory reduction reported below therefore reflects the detachment schedule alone. \texttt{E2E} is trained with a single global optimizer, whereas \texttt{s-DECOLLE} and \texttt{DELL} use one optimizer per block (eleven in total), preventing gradients from propagating across block boundaries. Peak training memory is defined as the maximum live-tensor footprint reached during a training pass. Each configuration is trained in a separate process on a single NVIDIA RTX A5000 GPU, so allocations from different runs never contribute to one another's peak. Table~\ref{tab:sota} reports every entry on the official DSEC test benchmark: ours submitted to the evaluation server, the baselines as their authors submitted them. Every other result in this section, namely the accuracy and memory comparison of Table~\ref{tab:tradeoff} and the depth ablation of Table~\ref{tab:depth-ablation}, is measured on the validation split, whose labels we hold. Test and validation figures are never mixed within a table, and absolute values are never compared across the two splits.

\rhead{Comparison to the state of the art}\label{sec:sota}

The first comparison in Table~\ref{tab:sota} is between our own two configurations: \texttt{DELL} outperforms \texttt{E2E} on every metric of the official benchmark, lowering EPE by 14\% (1.941 to 1.670\,px), AE by 15\% (6.324 to 5.347$^\circ$) and the outlier rate by 25\% (14.63\% to 10.98\%). Block-wise training therefore does not trade accuracy for memory on held-out data; it improves both. The depth ablation below shows the same pattern, with \texttt{E2E} degrading as capacity grows while \texttt{DELL} does not.

Against the published SNN baselines, \texttt{DELL} ranks second in both EPE and AE (Table~\ref{tab:sota}). It improves on Cuadrado et al.~\cite{cuadrado2023} by 2\% in EPE and 16\% in AE, while trailing SDformerFlow~\cite{tian2024sdformerflow} by only 4\% in EPE, 10\% in AE, and 9\% in outlier rate, despite using roughly 24$\times$ fewer parameters (2.3\,M vs.\ $\sim$54.9\,M). Our compact backbone therefore remains competitive with substantially larger SNNs, and provides a lightweight substrate for evaluating the learning rule. The outlier rate is the only metric on which \texttt{DELL} does not outperform both baselines (10.98\% vs. 10.31\% for Cuadrado et al. and 10.05\% for SDformerFlow), despite its lower EPE than Cuadrado et al.

Our network predicts flow at a fraction of the benchmark's resolution: a $192{\times}256$ grid, 2.5$\times$ smaller along each axis than DSEC's native $480{\times}640$, and the metrics of Table~\ref{tab:sota} are computed after bilinearly upsampling the predictions and rescaling their magnitude by $2.5\times$ to match the native grid (Setup above). This lower prediction resolution inevitably limits the recovery of fine spatial details and may affect metrics sensitive to local errors, while the SNN baselines of Table~\ref{tab:sota} predict directly at the native resolution. Despite this disadvantage, \texttt{DELL} remains close to SDformerFlow while using roughly 24$\times$ fewer parameters, making this a favorable operating point for a compact SNN.

For reference, E-RAFT~\cite{gehrig2021eraft}, which introduced the DSEC optical-flow benchmark and provides a strong non-spiking baseline with global learning, achieves a substantially lower EPE (0.779), but at significantly higher energy cost than our fully spiking approach; we revisit this comparison from an energy perspective in Section~\ref{sec:energy}.

\begin{table}[t]
\centering
\caption{\textbf{Optical flow on the official DSEC test benchmark}. Outliers (Out.) are pixels with EPE${}>{}$3\,px. Train.\ is end-to-end BPTT or our block-wise scheme (B-W, Section~\ref{sec:blockwise}). Every row is a submission to the DSEC evaluation server. Bold: best; underline: second best.}
\label{tab:sota}
\footnotesize
\setlength{\tabcolsep}{2pt}
\begin{tabular}{lccccc}
\toprule
Model & Train. & EPE (px) $\downarrow$ & AE ($^\circ$) $\downarrow$ & Out.\ (\%) $\downarrow$ & Par.\ (M) $\downarrow$ \\
\midrule
Cuadrado et al.~\cite{cuadrado2023} & BPTT & 1.707 & 6.338 & \underline{10.31\%} & \textbf{1.2} \\
SDformerFlow~\cite{tian2024sdformerflow} & BPTT & \textbf{1.602} & \textbf{4.871} & \textbf{10.05\%} & $\sim$54.9 \\
Ours (\texttt{E2E}) & BPTT & 1.941 & 6.324 & 14.63\% & \underline{2.3} \\
Ours (\texttt{DELL}) & B-W & \underline{1.670} & \underline{5.347} & 10.98\% & \underline{2.3} \\
\bottomrule
\end{tabular}
\end{table}

\rhead{Memory and accuracy of local learning}\label{sec:tradeoff}

\begin{figure*}[t]
\vspace*{3mm}
\centering
\setlength{\qualw}{0.121\textwidth}
\setlength{\tabcolsep}{0.5pt}
\renewcommand{\arraystretch}{0.4}
\begin{tabular}{cccccccc}
\qhead{Events} & \qhead{Ground truth} & \multicolumn{2}{c}{\qhead{\textbf{\texttt{DELL}} (ours)}} & \multicolumn{2}{c}{\qhead{\texttt{s-DECOLLE}}} & \multicolumn{2}{c}{\qhead{\texttt{E2E}}} \\
 & & \qhead{full} & \qhead{masked} & \qhead{full} & \qhead{masked} & \qhead{full} & \qhead{masked} \\
\qpanel{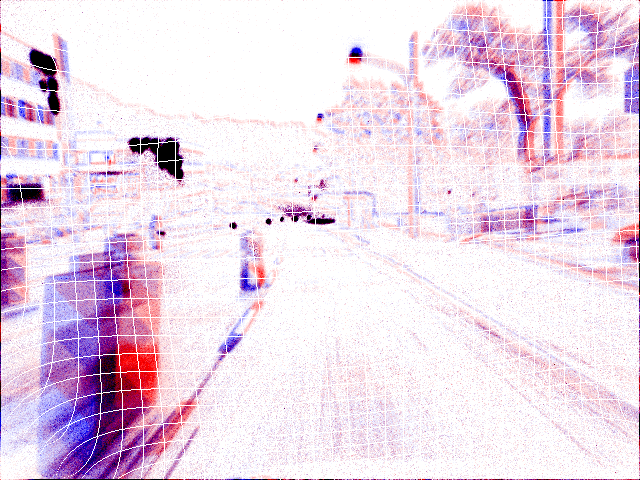} & \qpanel{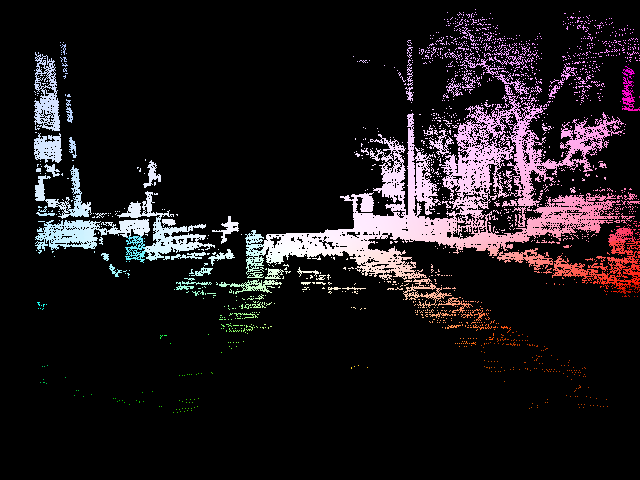} & \qpanel{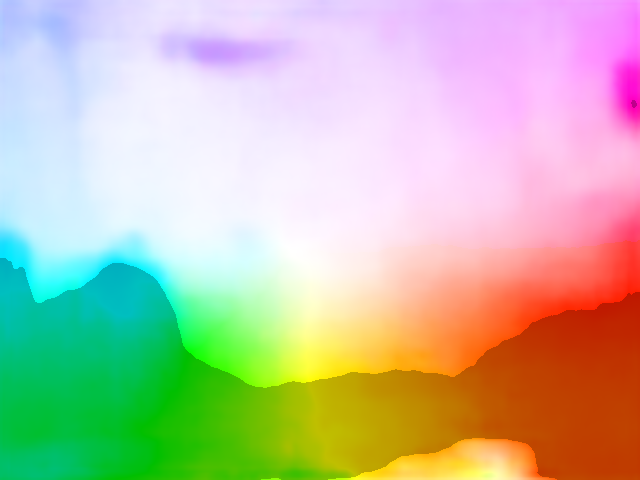} & \qpanel{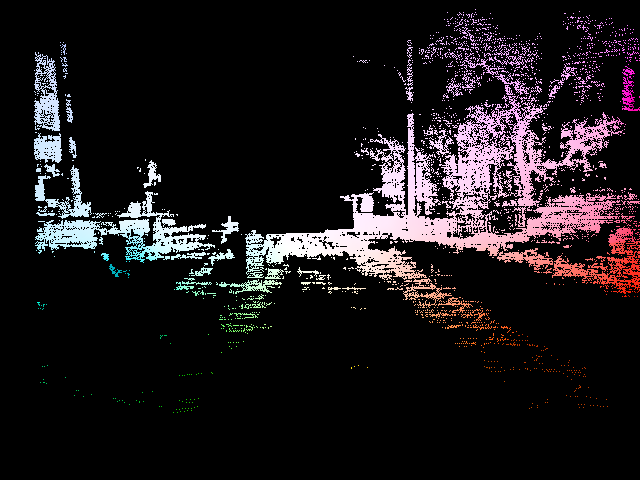} & \qpanel{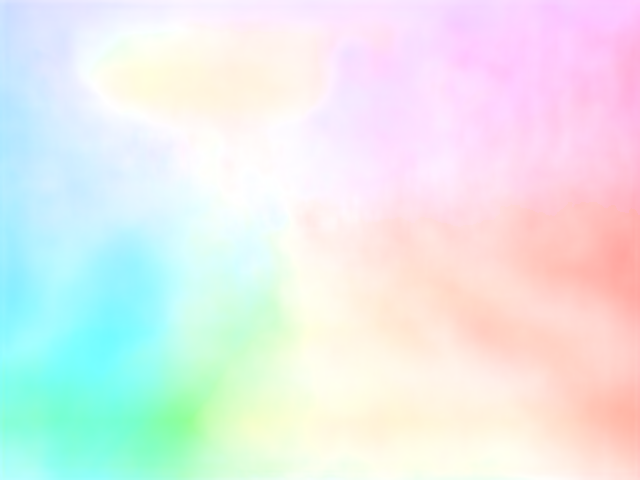} & \qpanel{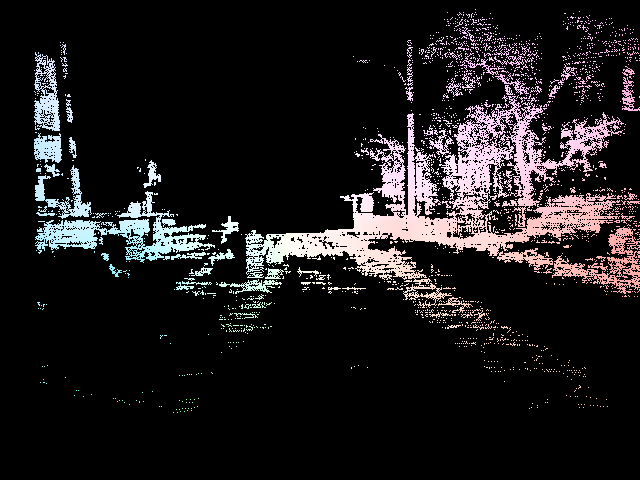} & \qpanel{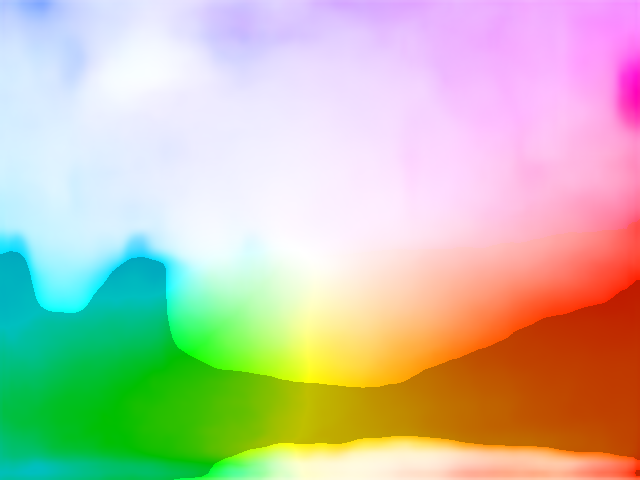} & \qpanel{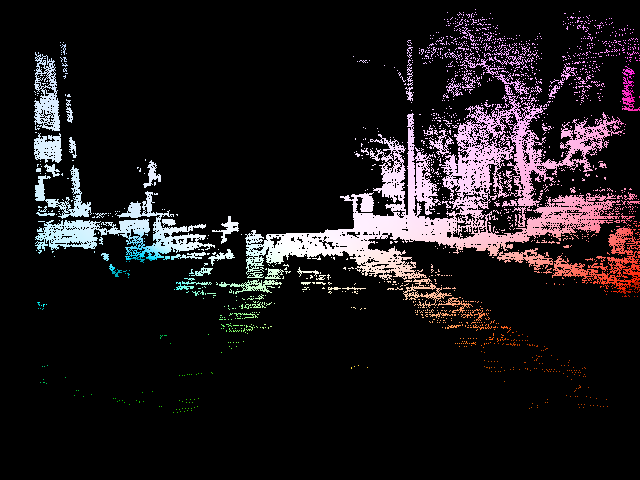} \\
\qpanel{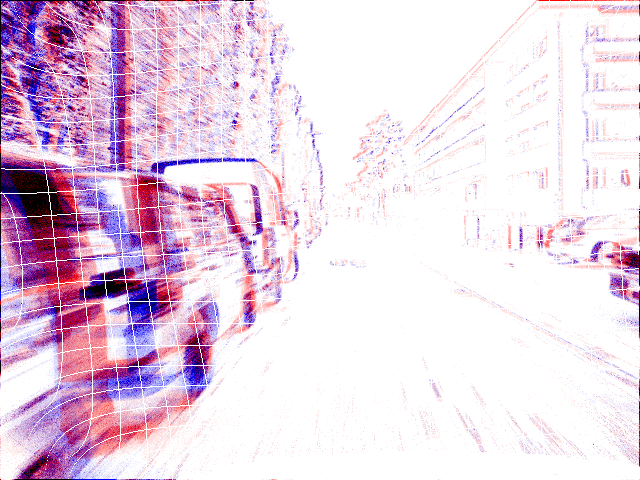} & \qpanel{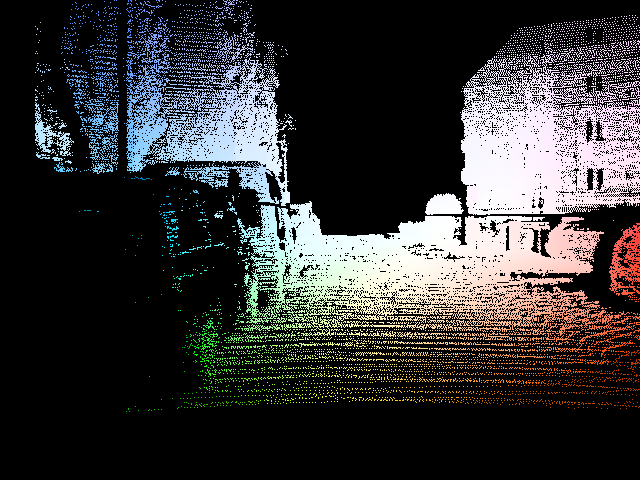} & \qpanel{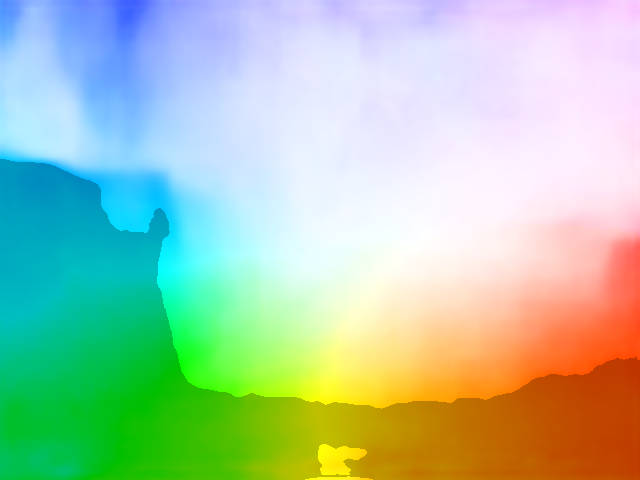} & \qpanel{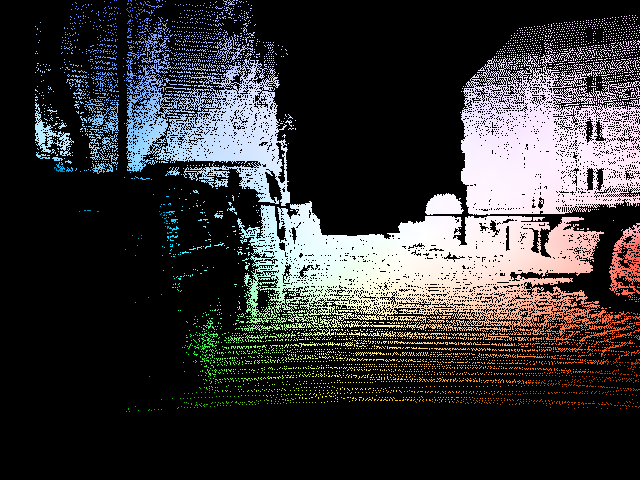} & \qpanel{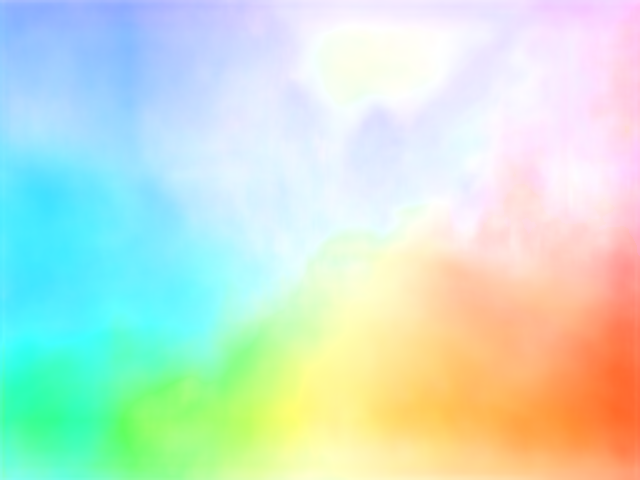} & \qpanel{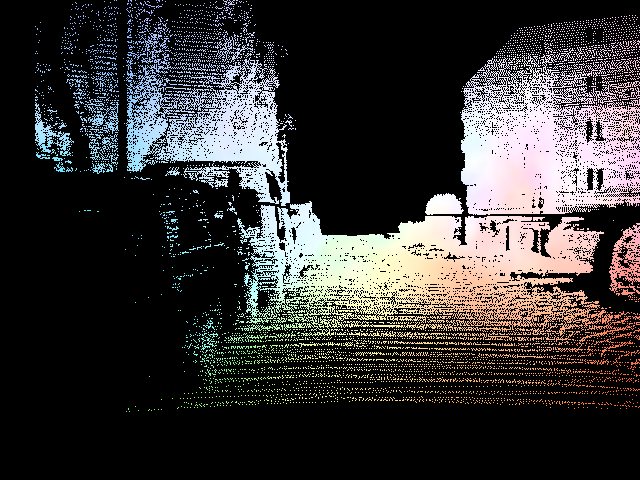} & \qpanel{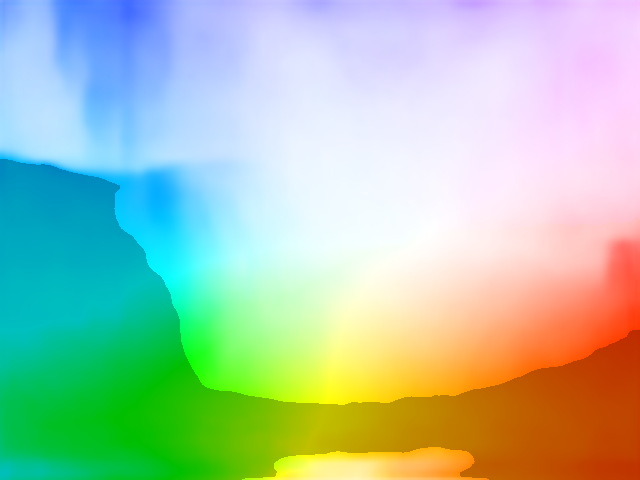} & \qpanel{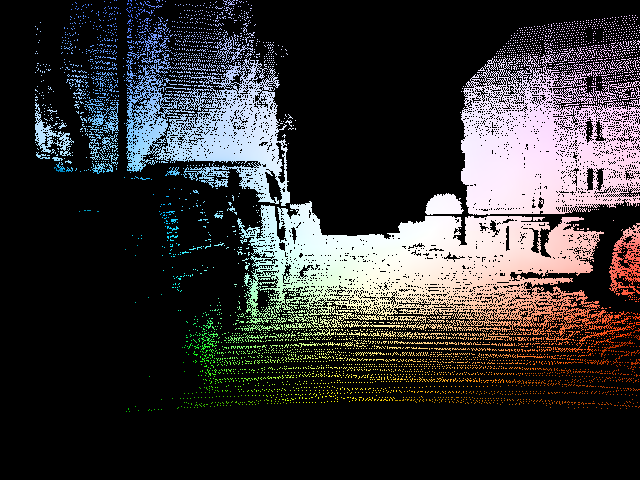} \\
\qpanel{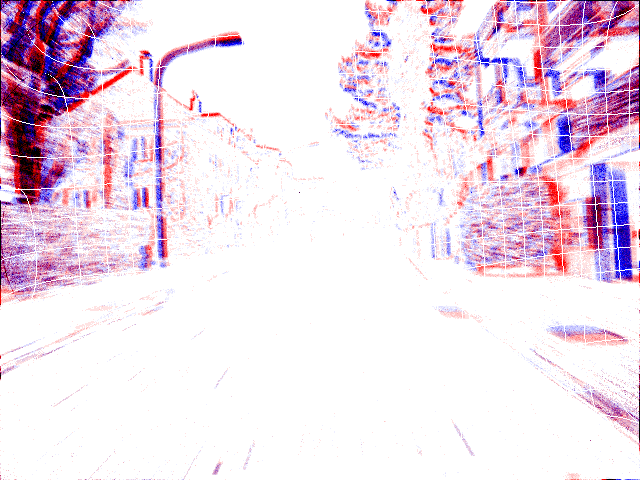} & \qpanel{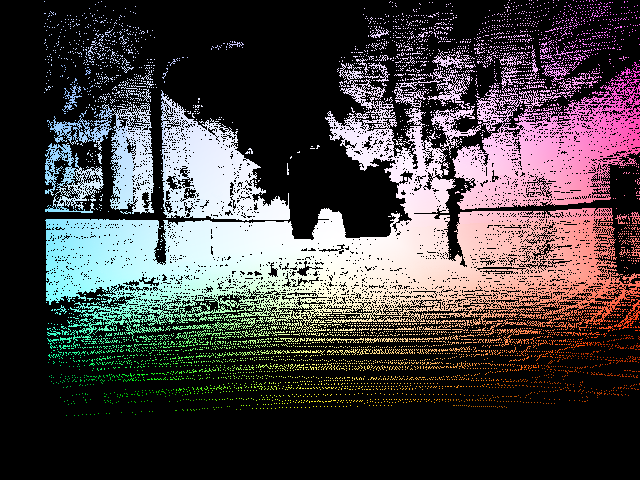} & \qpanel{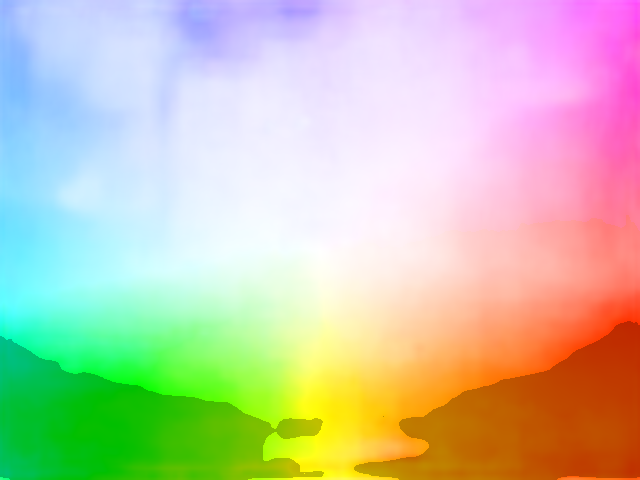} & \qpanel{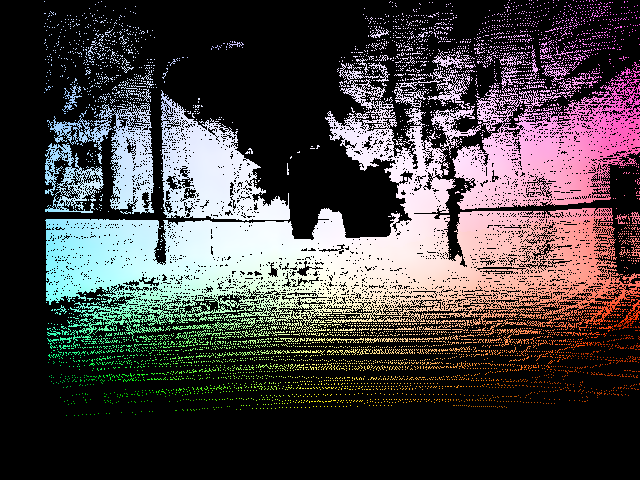} & \qpanel{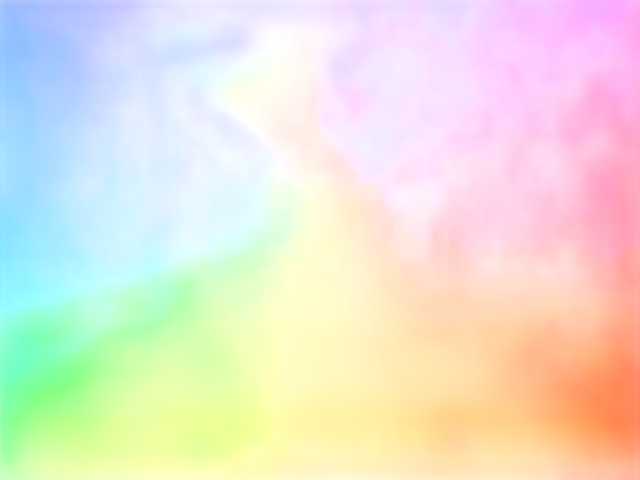} & \qpanel{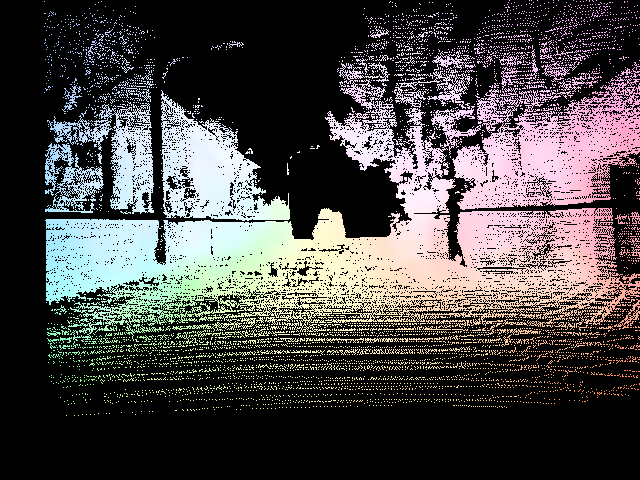} & \qpanel{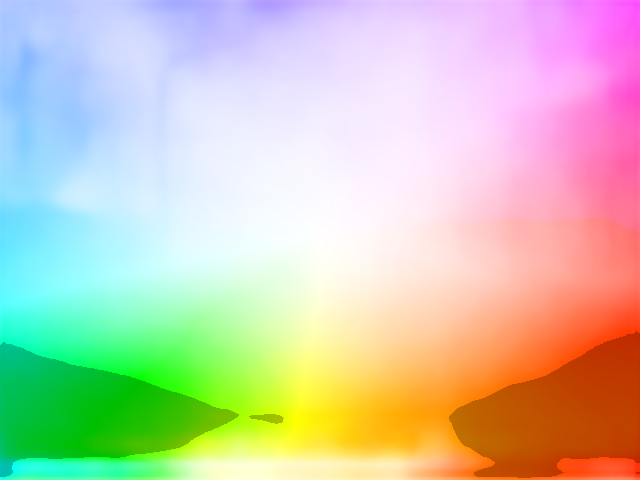} & \qpanel{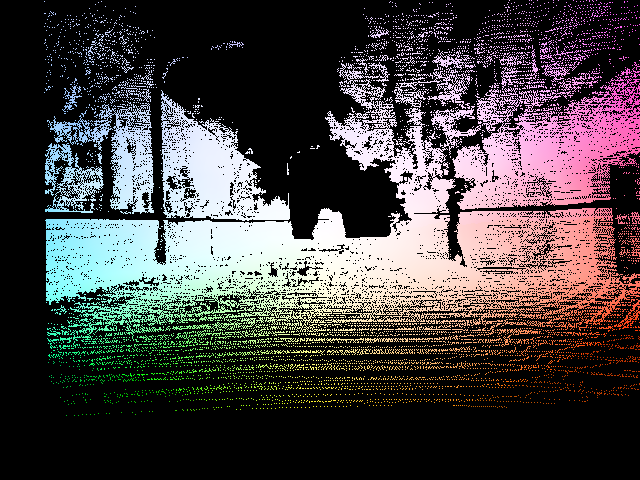} \\
\qhead{(a)} & \qhead{(b)} & \qhead{(c)} & \qhead{(d)} & \qhead{(e)} & \qhead{(f)} & \qhead{(g)} & \qhead{(h)} \\
\end{tabular}
\caption{Qualitative optical flow on DSEC. Each row is one validation sample: (a) the event input, (b) the ground truth, (c, d) \texttt{DELL} (block-wise, ours), (e, f) \texttt{s-DECOLLE} (frozen random local heads), and (g, h) \texttt{E2E} (end-to-end BPTT). Flow is color-coded with the usual hue--magnitude wheel, hue giving direction and brightness magnitude. The DSEC ground truth is sparse, so pixels without a valid target stay black; the masked columns restrict each prediction to those same pixels, which is where the metrics of Table~\ref{tab:tradeoff} are computed.}
\label{fig:qual-flow}
\end{figure*}

Block-wise local learning improves both terms (Table~\ref{tab:tradeoff}). Moving from \texttt{E2E} to \texttt{DELL} saves 39.6\% of peak training memory and lowers all three accuracy metrics on the validation split: EPE from 1.250 to 1.197\,px, AE from 5.86 to 5.54$^\circ$, and the outlier rate from 6.44\% to 5.42\%. The two gains have separate origins, the memory coming from the detachment schedule (Section~\ref{sec:blockwise}) and the accuracy from the regularizing effect that the depth ablation and the test benchmark both expose, so neither is paid for with the other. Accuracy is lost only when the local heads are frozen instead of learned: compared with \texttt{E2E}, \texttt{s-DECOLLE} exhibits 294\%, 144\%, and 651\% higher EPE, AE, and outlier rate, respectively, and its flow fields collapse to a near-uniform field (Figure~\ref{fig:qual-flow}). This contrasts with \texttt{DECOLLE}'s classification benchmarks, where fixed random projections suffice, although \texttt{s-DECOLLE} remains the most memory-efficient configuration. Learning the heads against a finer local target closes all three gaps and ends ahead of \texttt{E2E}, by 0.053\,px, 0.32$^\circ$ and 1.02 points, at a 29\% memory premium over \texttt{s-DECOLLE}. The 0.053\,px by which \texttt{DELL} leads \texttt{E2E} here widens to 0.271\,px on the test benchmark of Table~\ref{tab:sota}.

\begin{table}[t]
\centering
\caption{\textbf{Accuracy and peak training memory on validation split DSEC optical flow}, at identical backbone and detachment schedule. \texttt{s-DECOLLE} freezes every local head at its random initialization, \texttt{DELL} learns them. Peak memory is measured at equal batch size; \emph{saved} is the reduction relative to the \texttt{E2E} reference (green). The test-benchmark figures are in Table~\ref{tab:sota}.}
\label{tab:tradeoff}
\footnotesize
\setlength{\tabcolsep}{2.5pt}
\begin{tabular}{lcccccc}
\toprule
 & EPE & AE & Out. & \multicolumn{2}{c}{Peak memory} & Par.\ (M) \\
\cmidrule(lr){5-6}
Config. & (px)\,$\downarrow$ & ($^\circ$)\,$\downarrow$ & (\%)\,$\downarrow$ & (MB)\,$\downarrow$ & saved\,$\uparrow$ & train\,$\to$\,inf. \\
\midrule
\texttt{E2E} & 1.250 & 5.86 & 6.44 & 13501 & \emph{ref.} & 2.3\,$\to$\,2.3 \\
\texttt{s-DECOLLE} & 4.923 & 14.30 & 48.38 & 6316 & \memsave{53.2\%} & 2.4\,$\to$\,2.3 \\
\texttt{DELL} & \textbf{1.197} & \textbf{5.54} & \textbf{5.42} & 8157 & \memsave{39.6\%} & 7.9\,$\to$\,2.3 \\
\bottomrule
\end{tabular}
\end{table}

\rhead{Ablation: backbone depth}\label{sec:ablation-depth}

The comparison above is measured at a single backbone depth (depth$=4$, 11 blocks), the configuration used throughout Section~\ref{sec:expflow}. Since more depth introduces more detachment boundaries, both the accuracy and memory savings of block-wise learning may scale with it. We test this with the same backbone family as in Section~\ref{sec:architecture} (base\_channels$=32$), sweeping depth $\in\{3,4,5\}$; each added stage adds an encoder--decoder pair and doubles the channel width, so the block count rises from 9 to 13 and the inference parameter count grows steeply (Table~\ref{tab:depth-ablation}).

We report EPE alone in Table~\ref{tab:depth-ablation}, as the primary metric for this comparison. \texttt{DELL}'s AE follows the same trend, 6.43$^\circ$ at depth 3, 5.54$^\circ$ at depth 4 and 5.52$^\circ$ at depth 5. Across this sweep, the memory reduction from block-wise detachment stays roughly stable with depth rather than growing with detachment boundaries (Table~\ref{tab:depth-ablation}). Accuracy favors \texttt{DELL} at every depth, and by a margin that widens sharply with it: 0.011, 0.053 and 0.259\,px at depths 3, 4 and 5. \texttt{E2E}'s EPE \emph{worsens} from depth 4 to depth 5 (1.250 $\to$ 1.424), consistent with increased overfitting as network capacity grows, whereas \texttt{DELL}'s EPE \emph{improves} monotonically (1.366 $\to$ 1.197 $\to$ 1.165 from depth 3 to depth 5). Block-wise detachment therefore appears to limit how much the network overfits as depth increases.

Local learning becomes more attractive as networks grow deeper: its memory saving holds while its accuracy advantage over end-to-end training widens. Only half of the initial hypothesis holds, then: the accuracy advantage grows with depth, the memory saving does not. The official test benchmark orders the two configurations the same way and more sharply (Table~\ref{tab:sota}): end-to-end training loses more in the move from validation to test (EPE $+55\%$) than block-wise training does ($+40\%$), as a regularizing effect of detachment would predict.

\begin{table}[t]
\centering
\caption{\textbf{Backbone depth ablation: EPE (px) on the validation split},
\texttt{E2E} against \texttt{DELL}, with the inference parameter count
of each backbone and the peak-memory reduction of block-wise training
over \texttt{E2E}. Gap is \texttt{E2E} minus \texttt{DELL}; it is
positive at every depth, meaning \texttt{DELL} is the more accurate
(green).}
\label{tab:depth-ablation}
\footnotesize
\setlength{\tabcolsep}{2pt}
\begin{tabular}{lccccc}
\toprule
Depth & Par.\ (M) & \texttt{E2E} $\downarrow$ & \texttt{DELL} $\downarrow$ & Gap & Mem.\ saved $\uparrow$ \\
\midrule
3 & 0.8 & 1.377 & 1.366 & \memsave{0.011} & 39.3\% \\
4 (main paper) & 2.3 & 1.250 & 1.197 & \memsave{0.053} & 39.6\% \\
5 & 8.7 & 1.424 & \textbf{1.165} & \memsave{0.259} & 37.8\% \\
\bottomrule
\end{tabular}
\end{table}

\subsection{Semantic Segmentation}
\label{sec:segmentation}

We next consider semantic segmentation, a per-pixel \emph{classification} task, and ask whether the accuracy results of Section~\ref{sec:expflow} generalize to it. We evaluate two datasets: DSEC-Semantic~\cite{sun2022ess}, which enables direct comparison with the SNN state of the art, and M3ED~\cite{chaney2023m3ed}.

\rhead{Setup}\label{sec:segmentation-setup}

DSEC-Semantic~\cite{sun2022ess} annotates the same recordings as DSEC~\cite{gehrig2021dsec} (Section~\ref{sec:expflow}) under the reduced 11-class ESS / DSEC-Semantic scheme~\cite{sun2022ess}, the same convention SpikingEDN~\cite{zhang2024spikingedn} reports against. M3ED~\cite{chaney2023m3ed} is instead labeled by the recent M3ED-Semantic annotations of Kong et al.~\cite{kong2025eventfly}; we adopt the same 11-class scheme on M3ED for consistency rather than a bespoke taxonomy, keep the $T{=}16$ binning of Section~\ref{sec:expflow}, and downsample to $144{\times}256$. M3ED-Semantic defines no standard split, so we use a custom 80/20 train/validation partition over the in-domain vehicle sequences, released with our code.

The backbone and the three training configurations (Sections~\ref{sec:architecture} and~\ref{sec:blockwise}) carry over from the optical-flow study. The read-out and each local head are altered, from a 2-channel flow output to an 11-channel per-pixel class-logit output. Every configuration's output is the read-out prediction alone, and \texttt{E2E} is trained with the read-out loss alone (Section~\ref{sec:blockwise}). The SmoothL1 segmentation loss replaces the modulus-angular flow loss (Section~\ref{sec:blockwise}). On DSEC-Semantic we report mean intersection-over-union (mIoU) alone, to match SpikingEDN~\cite{zhang2024spikingedn}; on M3ED, where no baseline constrains the comparison, we report pixel accuracy (Acc), mean class accuracy (mAcc), mIoU, and frequency-weighted IoU (fIoU).

\rhead{DSEC-Semantic}\label{sec:segmentation-results} As discussed in Section~\ref{sec:related}, SpikingEDN~\cite{zhang2024spikingedn} reports 52.71\% mIoU on DSEC-Semantic in an events-only configuration comparable to ours. EvSegSNN~\cite{hareb2024evsegsnn}, in contrast, reports results only on DDD17 and does not appear in Table~\ref{tab:seg-dsec}. DSEC-Semantic thus provides the head-to-head comparison with the SNN state of the art that M3ED alone cannot offer, as Table~\ref{tab:sota} does for optical flow.

\texttt{E2E} reaches 48.4\% mIoU and \texttt{DELL} 44.7\%, against SpikingEDN's 52.71\% (Table~\ref{tab:seg-dsec}). We trail the state of the art by 4.3 and 8.0 points respectively, with 3.7$\times$ fewer inference parameters (2.3M against 8.5M).

The contrast with optical flow is informative. On DSEC flow, \texttt{s-DECOLLE} performs substantially worse: its EPE is 3.9$\times$ that of \texttt{E2E}, while \texttt{DELL}'s learned heads more than recover that gap (Section~\ref{sec:expflow}). On semantic segmentation, by contrast, \texttt{s-DECOLLE} reaches 40.7\% mIoU, only 7.7 percentage points below \texttt{E2E}, and learned heads recover 52\% of this gap. Fixed random local read-outs degrade gracefully when the local target is a bounded per-pixel class distribution, and fail when it is an unbounded continuous vector field. What \texttt{DECOLLE}'s frozen heads lack is not dense supervision but the capacity to reach an arbitrary output scale: a random projection can order 11 bounded class scores, but cannot match the magnitude of a flow vector it never sees.

\begin{table}[t]
\centering
\caption{\textbf{Semantic segmentation on DSEC-Semantic} (11 classes~\cite{sun2022ess}), against the SNN state of the art. SpikingEDN's events-only LIF configuration runs at $T{=}1$ against our $T{=}16$. Bold: best; underline: second best.}
\label{tab:seg-dsec}
\small
\setlength{\tabcolsep}{4pt}
\begin{tabular}{lcc}
\toprule
Model & Params (M) $\downarrow$ & mIoU $\uparrow$ \\
\midrule
SpikingEDN (events only)~\cite{zhang2024spikingedn} & 8.5 & \textbf{52.71\%} \\
Ours (\texttt{E2E}) & \textbf{2.3} & \underline{48.4\%} \\
Ours (\texttt{s-DECOLLE}) & \textbf{2.3} & 40.7\% \\
Ours (\texttt{DELL}) & \textbf{2.3} & 44.7\% \\
\bottomrule
\end{tabular}
\end{table}

\rhead{M3ED} No prior SNN method has been evaluated on M3ED (Section~\ref{sec:related}), and its semantic labels~\cite{kong2025eventfly} are recent. SpikingEDN~\cite{zhang2024spikingedn}, our reference on DSEC-Semantic, has no public implementation and cannot be retrained here, so we present these results as a reference point for future SNN methods on M3ED rather than as a competitive claim. \texttt{DELL} is slightly ahead of \texttt{E2E} on pixel accuracy and frequency-weighted IoU, and behind it on the two class-averaged metrics, so its deficit is concentrated in the rare and small classes rather than spread over the whole map. Since every configuration's output is the read-out prediction alone (Section~\ref{sec:blockwise}), this does not come from output fusion; a likely cause is that the coarser blocks are trained against pooled soft targets, which dilute small and thin classes. \texttt{s-DECOLLE} trails both by 3.6 to 6.4 points depending on the metric, and its 6.1-point mIoU deficit relatively matches the 7.7 points it gives up on DSEC-Semantic, so freezing the local heads costs about the same on either dataset.

\begin{table}[t]
\centering
\caption{\textbf{Semantic segmentation on M3ED} (11 classes~\cite{sun2022ess}), the first SNN results on this benchmark.}
\label{tab:seg-m3ed}
\small
\setlength{\tabcolsep}{3pt}
\begin{tabular}{lcccc}
\toprule
Model & Acc $\uparrow$ & mAcc $\uparrow$ & mIoU $\uparrow$ & fIoU $\uparrow$ \\
\midrule
Ours (\texttt{E2E}) & 81.0 & 47.9 & 39.0 & 69.2 \\
Ours (\texttt{s-DECOLLE}) & 77.4 & 41.5 & 32.9 & 64.1 \\
Ours (\texttt{DELL}) & 81.2 & 47.0 & 38.3 & 69.4 \\
\bottomrule
\end{tabular}
\end{table}

\begin{figure}[t]
\centering
\setlength{\qualw}{0.238\columnwidth}
\setlength{\tabcolsep}{0.5pt}
\renewcommand{\arraystretch}{0.4}
\begin{tabular}{cccc}
\qhead{Events} & \qhead{\textbf{\texttt{DELL}} (ours)} & \qhead{\texttt{E2E}} & \qhead{Ground truth} \\
\qpanel{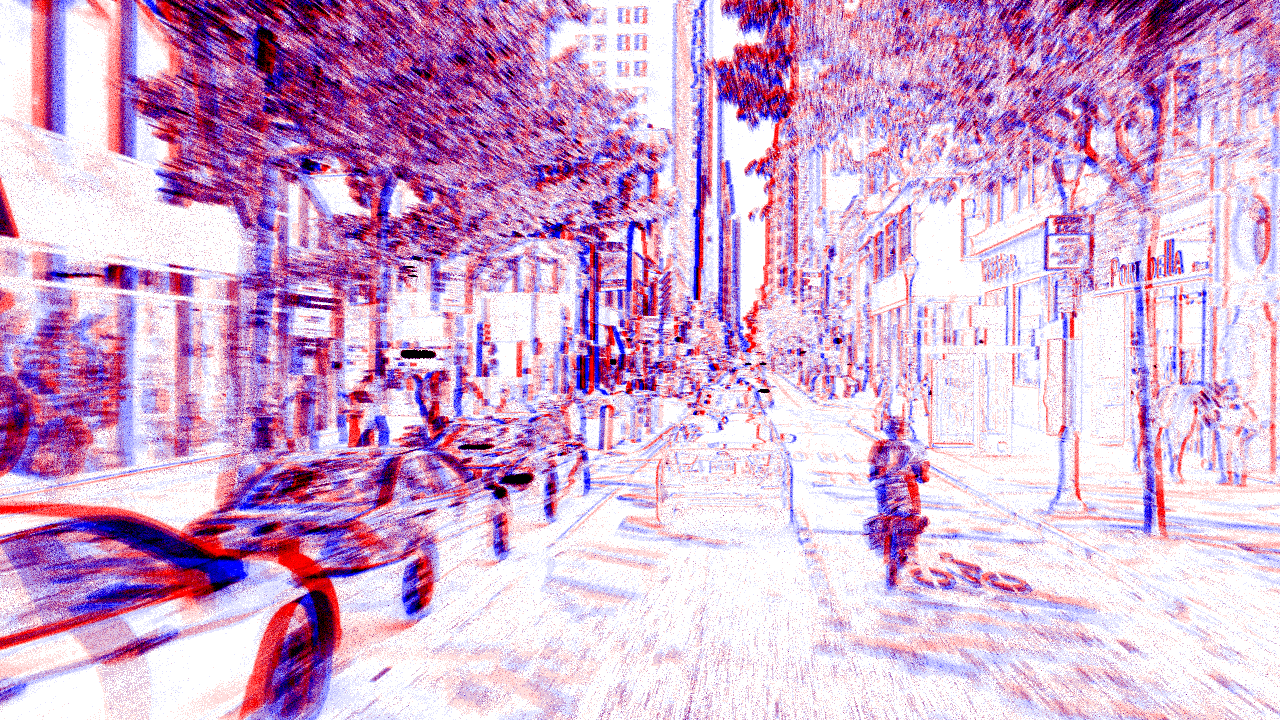} & \qpanel{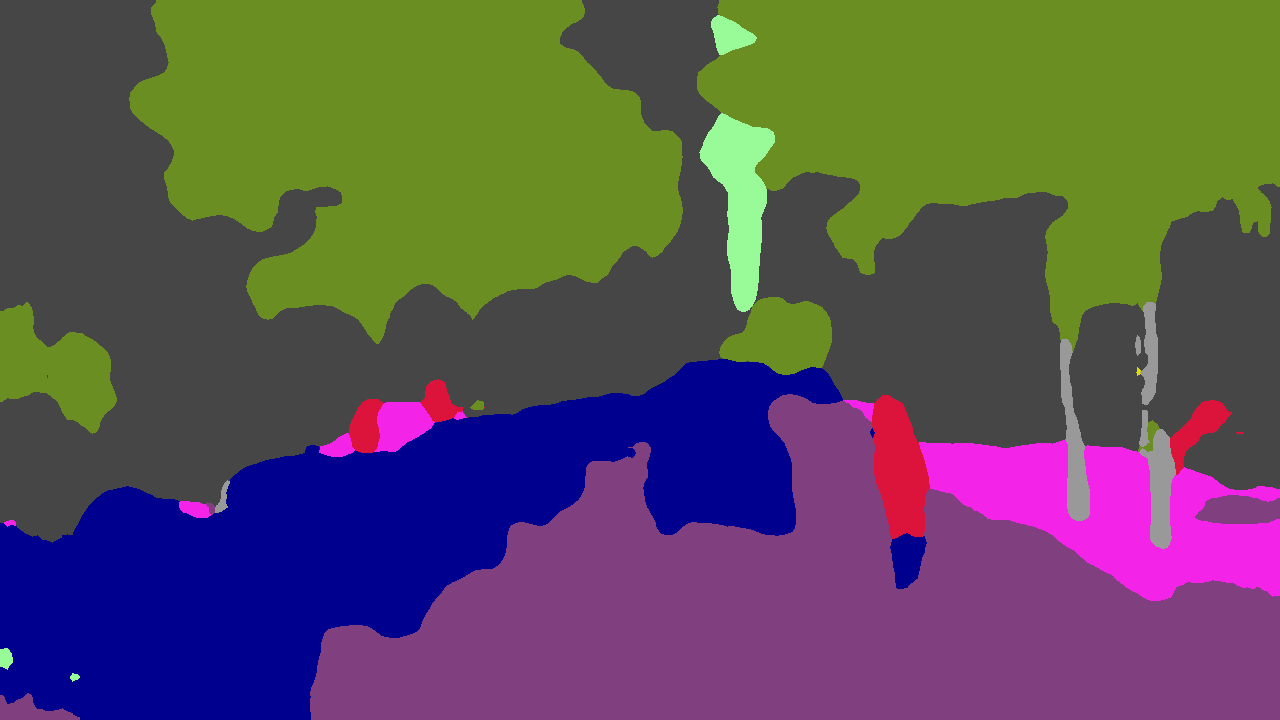} & \qpanel{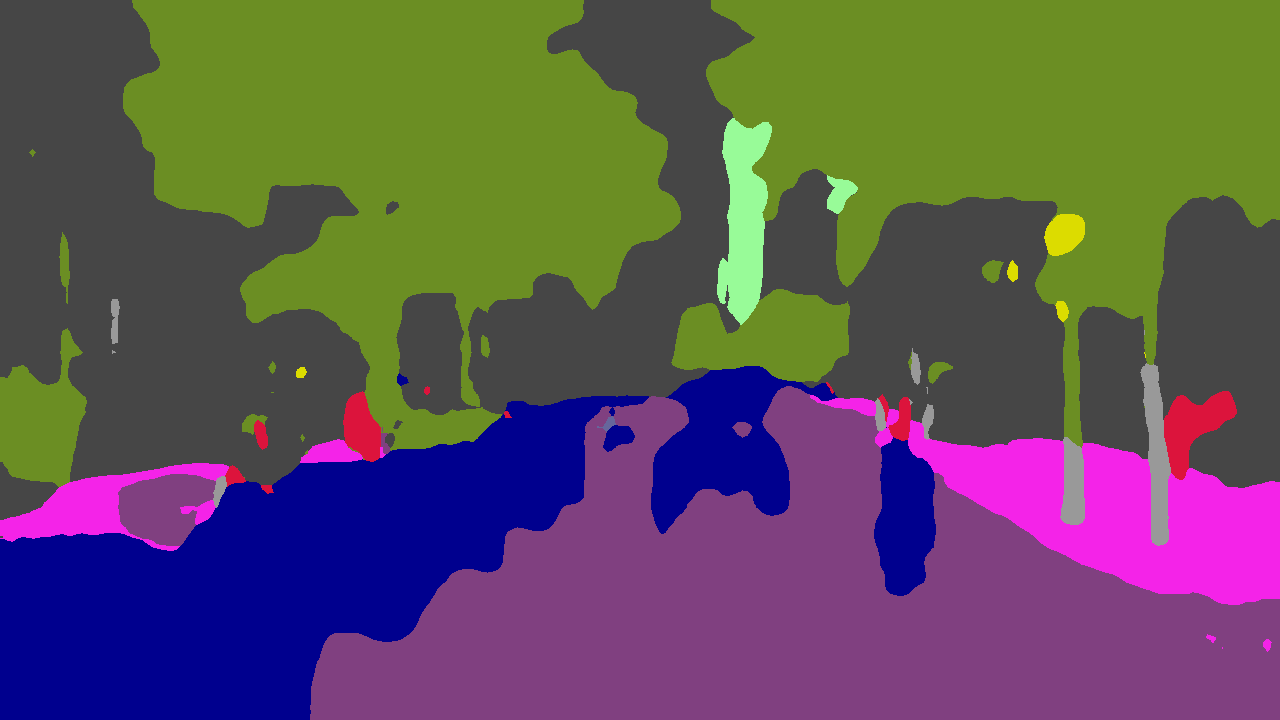} & \qpanel{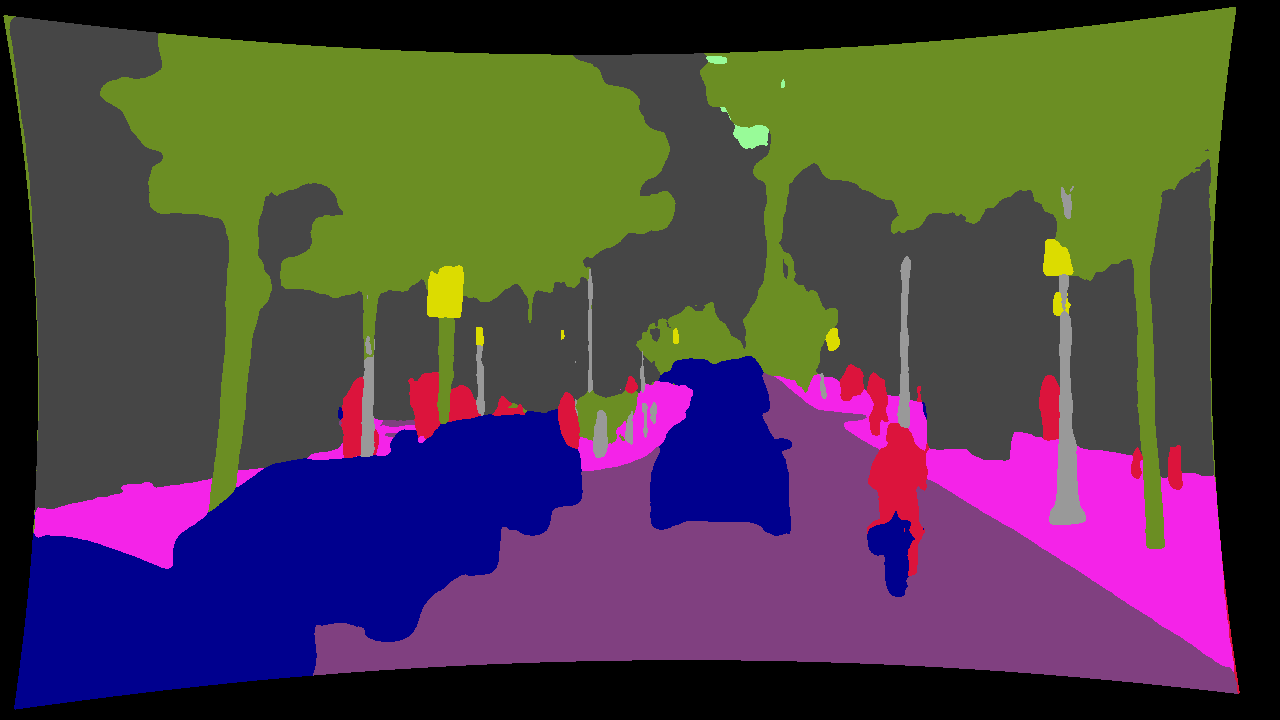} \\
\qpanel{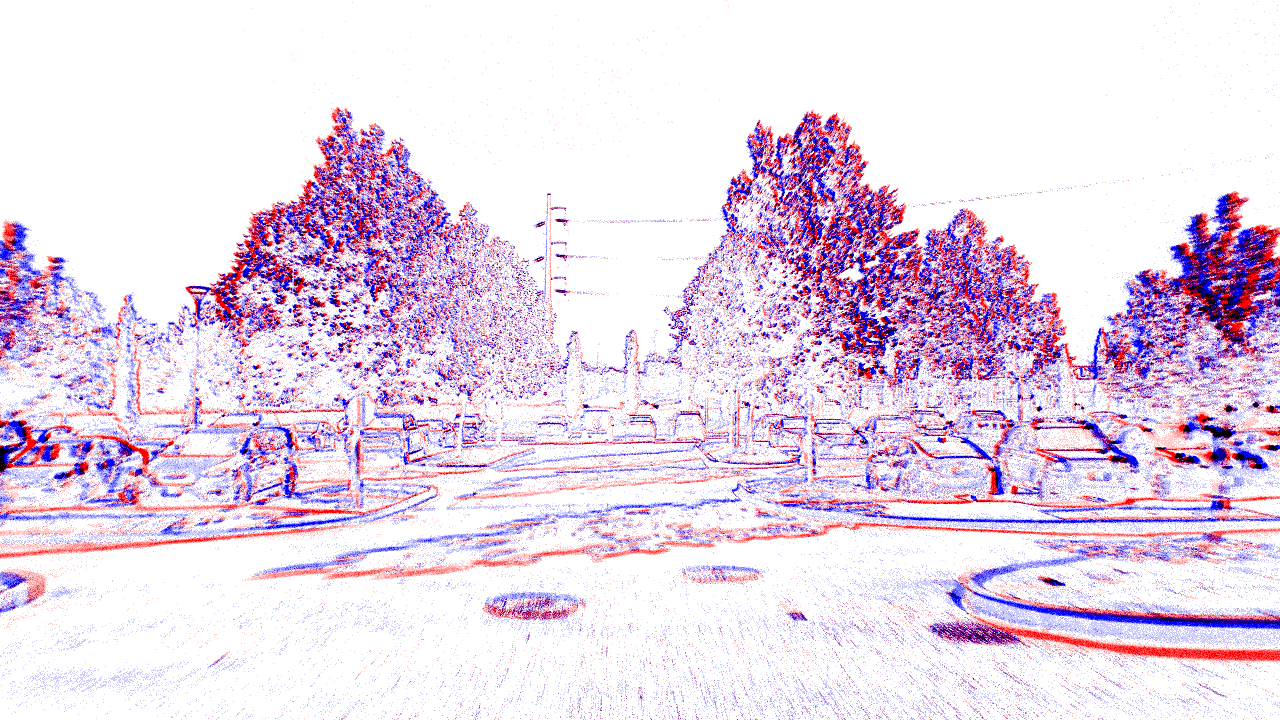} & \qpanel{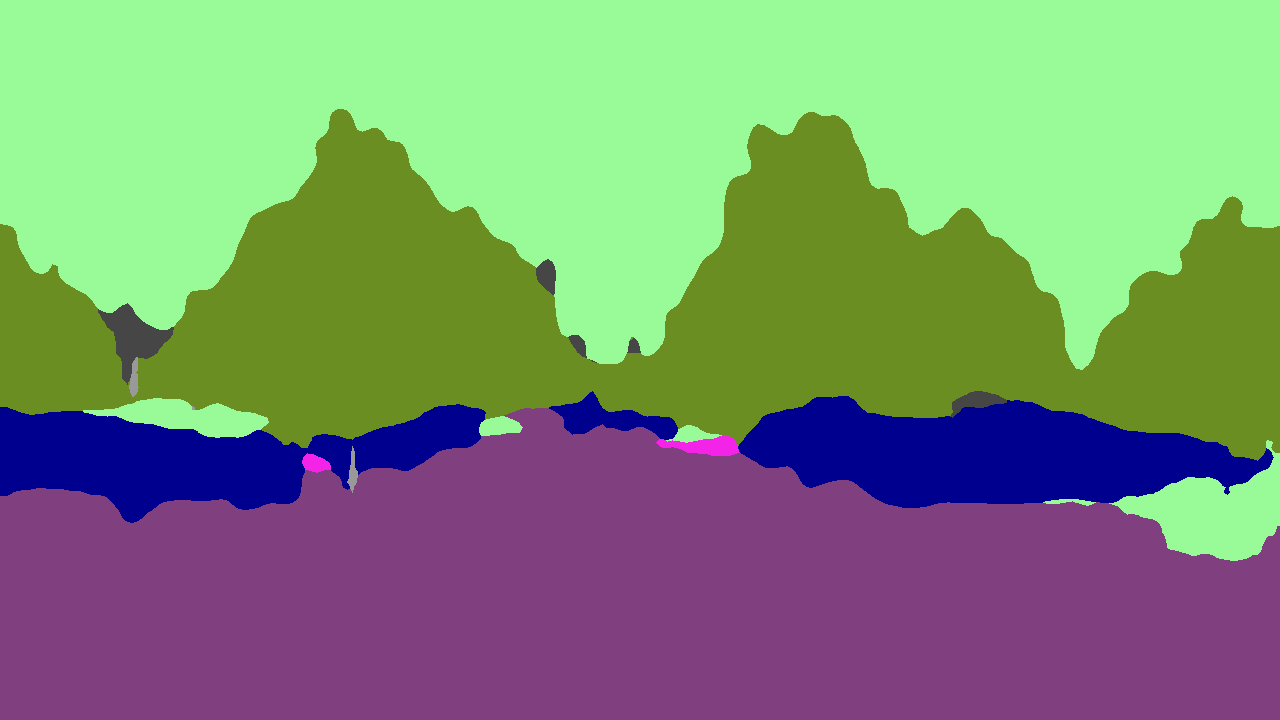} & \qpanel{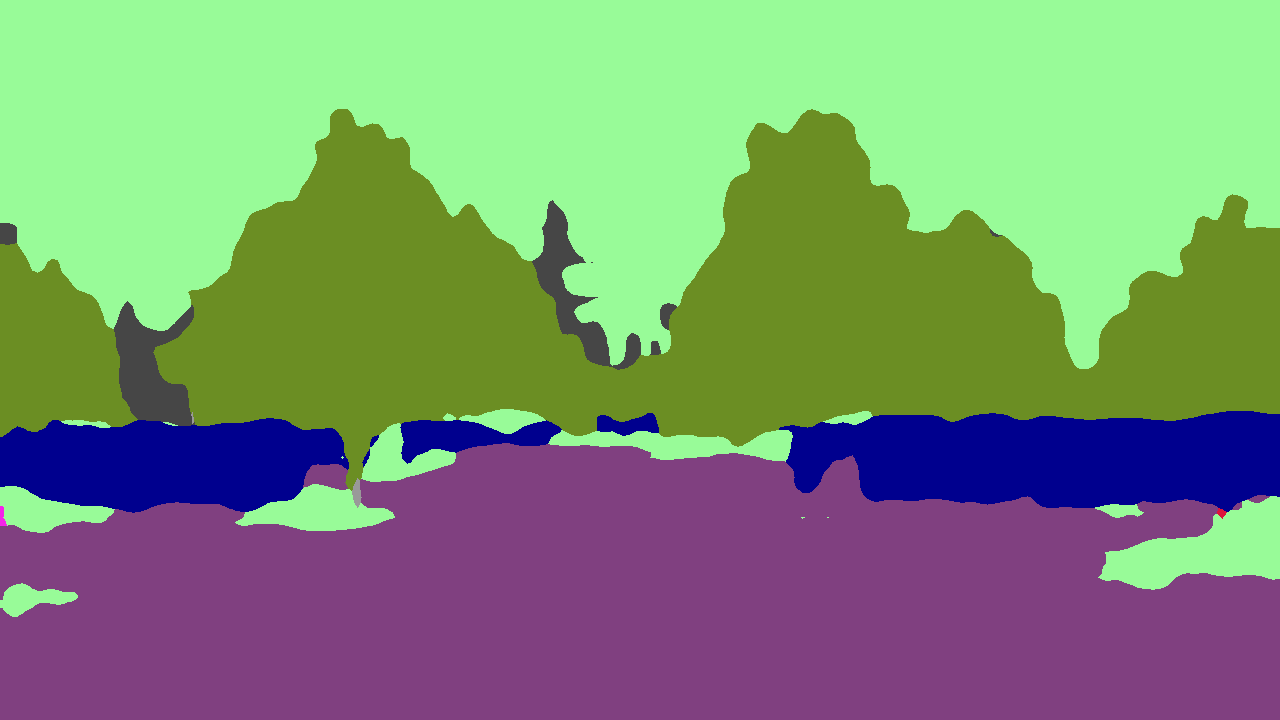} & \qpanel{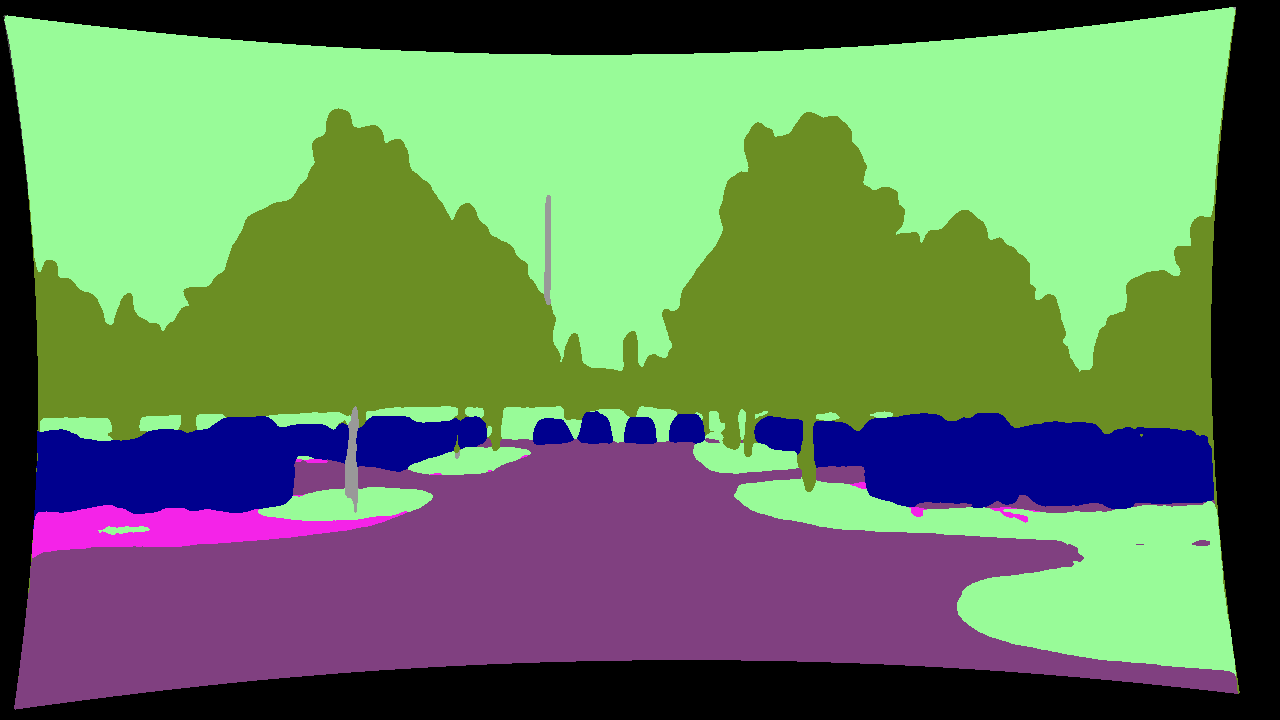} \\
\qpanel{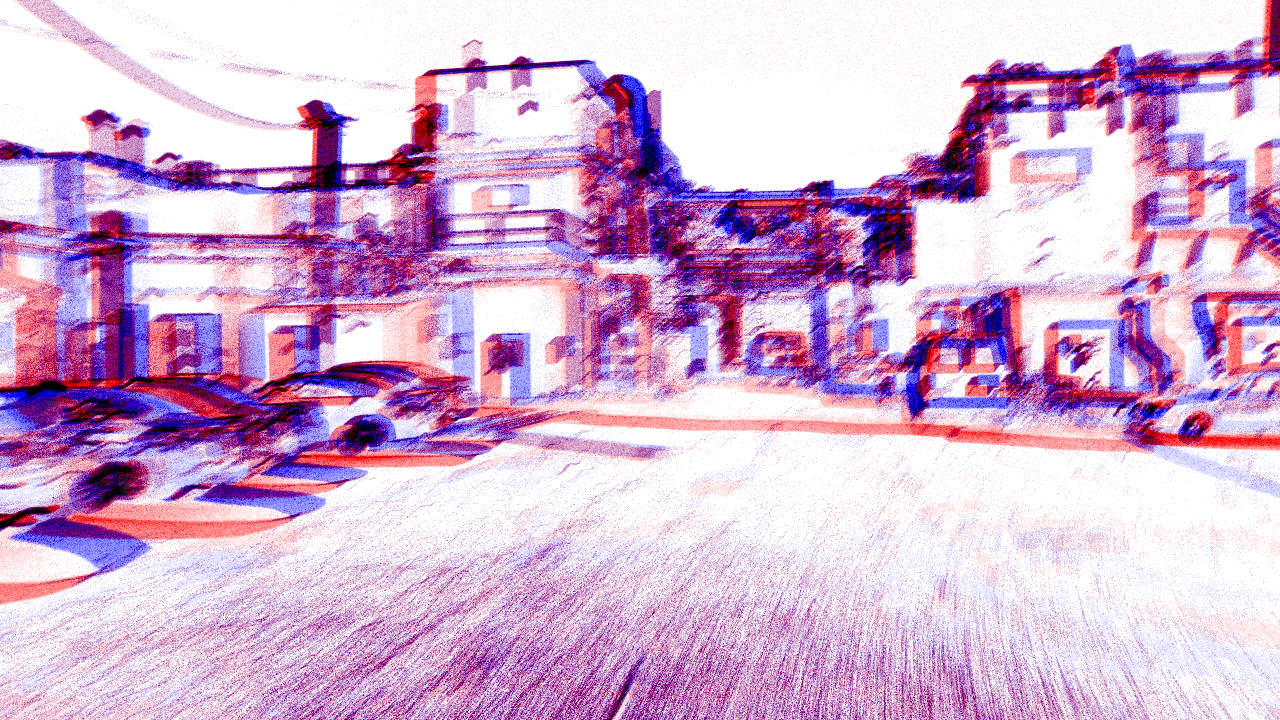} & \qpanel{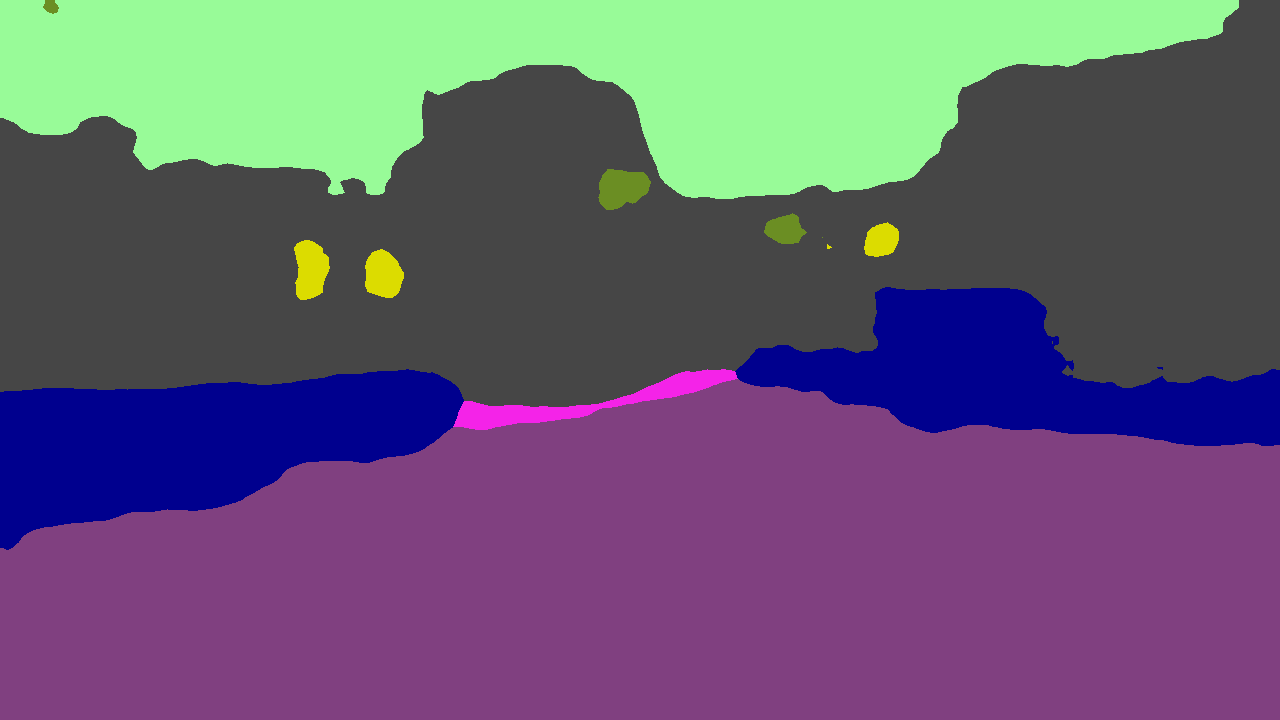} & \qpanel{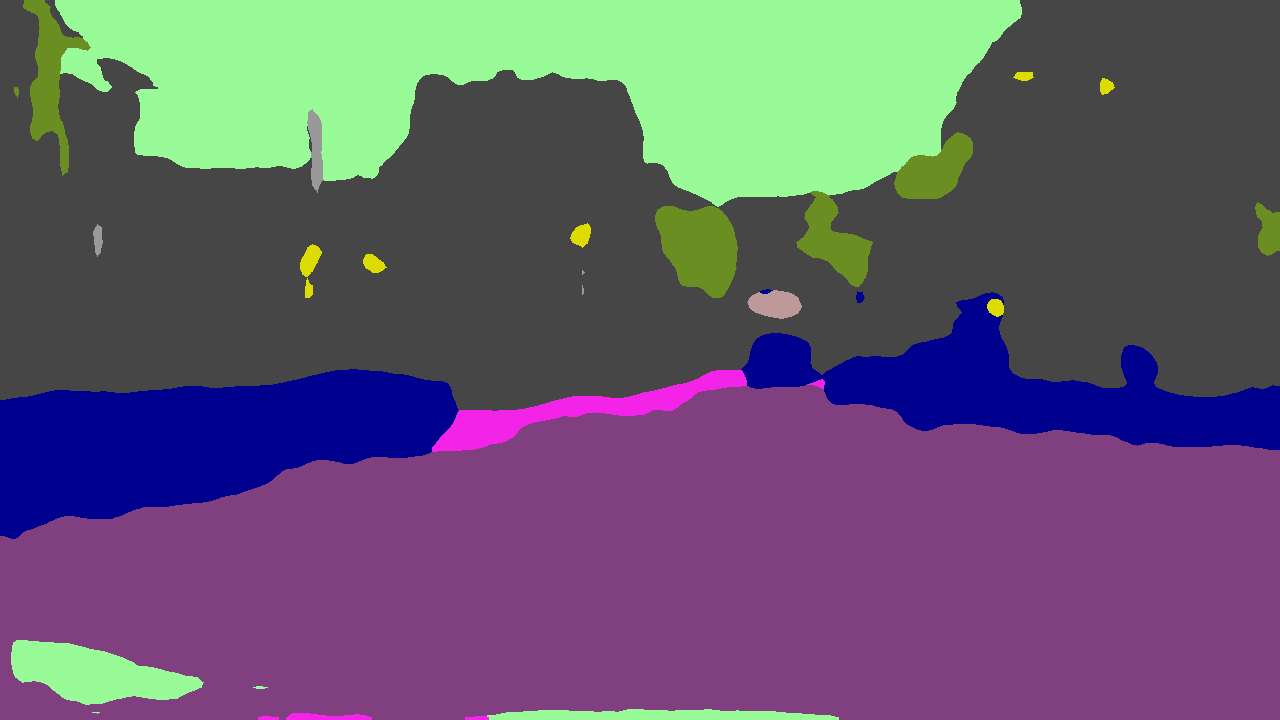} & \qpanel{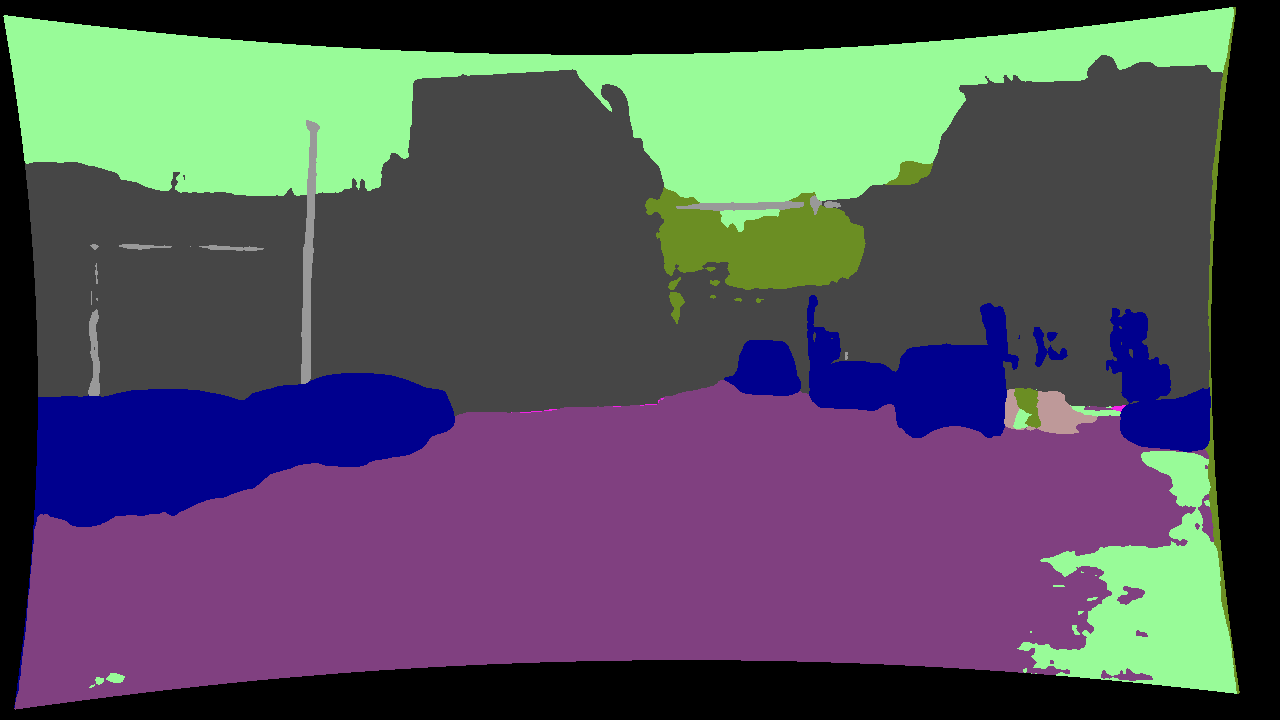} \\
\qhead{(a)} & \qhead{(b)} & \qhead{(c)} & \qhead{(d)} \\
\end{tabular}
\caption{Qualitative semantic segmentation on M3ED, in the style of Figure~\ref{fig:qual-flow}: (a) the event input, (b) \texttt{DELL} (block-wise, ours), (c) \texttt{E2E} (end-to-end BPTT), and (d) the ground truth, in the 11-class scheme~\cite{sun2022ess}.}
\label{fig:qual-seg}
\end{figure}

\subsection{Theoretical Energy Consumption}
\label{sec:energy}

We estimate the energy of a single inference analytically, from operation counts derived from the architecture rather than from a profiler. We follow the layer-wise accounting of Rueckauer et al.~\cite{rueckauer2017conversion}, as applied to event-based optical flow by Kosta and Roy~\cite{kosta2023adaptivespikenet}: a non-spiking layer costs one multiply-accumulate (MAC) per synaptic connection per forward pass, while a spiking layer costs one accumulate (AC) per connection only when the presynaptic neuron fires, repeated at every timestep, with the depthwise and pointwise stages of a separable convolution counted separately. We take a uniform firing rate of 0.1 and $T{=}16$ (Section~\ref{sec:expflow}). What differs from that accounting is the price of an operation, which we take from fabricated accelerators rather than from generic CMOS arithmetic energies: Loihi~1~\cite{davies2018loihi}, at 23.6\,pJ per synaptic operation, for the spiking networks, and Eyeriss~v1~\cite{chen2017eyeriss}, at $\sim$20\,pJ per MAC, for the non-spiking baseline. Cuadrado et al.~\cite{cuadrado2023} report no energy figures, so their row is our own count of their published architecture; their Conv3d temporal-fusion path is not directly mappable to Loihi, and we count its binary-input layers as spike-gated like the rest.

\begin{table}[t]
\centering
\caption{\textbf{Theoretical energy per inference on DSEC optical flow}. MACs for the non-spiking baseline, spike-gated ACs for the SNNs at a uniform 10\% firing rate, priced on the accelerator matching each architecture~\cite{davies2018loihi,chen2017eyeriss}; n/a marks the pairings that do not apply. Ours at $T{=}16$.}
\label{tab:energy}
\small
\setlength{\tabcolsep}{3pt}
\begin{tabular}{lccc}
\toprule
 & & \multicolumn{2}{c}{Energy (J)} \\
\cmidrule(lr){3-4}
Model & Ops / inf.\ $\downarrow$ & Loihi~1 & Eyeriss~v1 \\
\midrule
E-RAFT~\cite{gehrig2021eraft} (ANN) & 247.5\,GMAC & n/a & 4.950 \\
Cuadrado et al.~\cite{cuadrado2023} (SNN) & 41.0\,GAC & 0.967 & n/a \\
Ours (SNN) & \textbf{24.2\,GAC} & \textbf{0.571} & n/a \\
\bottomrule
\end{tabular}
\end{table}

The three configurations share the same inference network and the same estimated inference energy. Our backbone needs 24.2\,GAC per flow field, against 41.0\,GAC for Cuadrado et al.~\cite{cuadrado2023} and 247.5\,GMAC for E-RAFT~\cite{gehrig2021eraft}.

The operation count is the primary result and the joule values are derived from it, so these are order-of-magnitude estimates despite the precision at which they are tabulated. The 0.1 firing rate is an approximation rather than a measured statistic, so the comparison between the two SNNs reduces to a comparison of operation counts. The estimates are not normalized to the same input duration, our 100\,ms window being substantially longer than the ${\sim}9$\,ms used by Cuadrado et al. The 8.7$\times$ ratio to E-RAFT conflates the spiking paradigm with two accelerator families fabricated at different process nodes. Loihi~1's 23.6\,pJ is reported as a minimum rather than a workload average, making the spiking estimates optimistic by an unquantified amount. None of these estimates has been validated on silicon, and the analysis covers optical flow on DSEC alone; the segmentation results of Section~\ref{sec:segmentation} are not covered by Table~\ref{tab:energy}.

\section{Discussion and Limitations}
\label{sec:discussion}

Our results show that block-wise local learning can extend beyond classification to dense event-based prediction, while reducing the memory required for training. The 39.6\% reduction in peak memory comes from removing the dependency across network depth, while the accuracy improvements observed on DSEC suggest that this detachment can also act as a useful regularizer. The same local-learning principle transfers to semantic segmentation, although without the accuracy gain over end-to-end training seen on optical flow, so the effectiveness of local heads depends on the prediction task.

However, several limitations remain before this approach can constitute on-chip training. All experiments were performed off-chip on GPUs in full FP32 precision, and the energy analysis of Section~\ref{sec:energy} is analytical rather than measured. Although the backbone avoids operations that are difficult to map to neuromorphic hardware, its mappability has not been validated on silicon. Moreover, BPTT is still required within each block, so temporal dependencies over the $T{=}16$ input window remain in memory. Finally, local targets are precomputed off-chip from the full-resolution ground truth. An on-chip implementation would therefore require a mechanism for generating or broadcasting local targets without relying on the complete target map. These limitations motivate future work on temporally local learning, hardware validation, and on-chip generation of local supervision.

\section{Conclusion}
\label{sec:conclusion}

Local learning has so far only been demonstrated for classification, where a single global label can supervise every block; dense per-pixel prediction offers no such label. We introduced \texttt{DELL}, a block-wise scheme giving each block its own dense local target through a learnable head, and evaluated it on optical-flow regression and semantic segmentation with a compact 2.3M-parameter fully spiking backbone, competitive with the SNN state of the art on the official DSEC benchmark at 24$\times$ fewer parameters than the strongest baseline. On DSEC optical flow, \texttt{DELL} trains with 39.6\% less peak memory than end-to-end BPTT and is also more accurate than it, by 0.053\,px of endpoint error on our validation split and by 0.271\,px on the test benchmark, whereas the fixed random read-outs of \texttt{DECOLLE} do not transfer to this task and learnable heads more than recover their degradation. The advantage of detachment grows with depth while the memory saving holds, favoring block-wise training as networks grow.
\section*{Acknowledgments}

This work is under the programme DesCartes and is supported by the National Research Foundation, Prime Minister's Office, Singapore, under its Campus for Research Excellence and Technological Enterprise (CREATE) programme. This work was also supported by the ANR grant ANR-11-LABX-0040 within the French State Programme ``Investissements d'Avenir''.

We declare the following use of LLMs in the preparation of this work. For writing, OpenAI's ChatGPT and Anthropic's Claude were used to improve the phrasing and the clarity of some sentences of this article. For coding, Anthropic's Claude was used to speed up development, in particular to factorize existing scripts and to generate the plots. No LLM was used to produce the scientific content, the experiments, or their analysis.

\bibliographystyle{IEEEtran.bst}
\bibliography{references}
\end{document}